%% file: main.tex
\documentclass[10pt,twocolumn,letterpaper]{article}

\usepackage[pagenumbers]{cvpr} % To force page numbers, e.g. for an arXiv version

\input{preamble}
\definecolor{cvprblue}{rgb}{0.21,0.49,0.74}
\usepackage[pagebackref,breaklinks,colorlinks,allcolors=cvprblue]{hyperref}

\usepackage{soul}
\usepackage{url}
\usepackage[utf8]{inputenc}
\usepackage{graphicx}
\usepackage{amsmath}
\usepackage{amsthm}
\usepackage{booktabs}
\usepackage{algorithm}
\usepackage{algorithmic}
\usepackage[switch]{lineno}
\usepackage{amssymb}
\usepackage{pifont}
\usepackage{xcolor}
\usepackage{amssymb}
\usepackage{colortbl}
\usepackage[normalem]{ulem}
\useunder{\uline}{\ul}{}
\usepackage[inline]{enumitem}
\usepackage{enumitem}
\definecolor{customgreen}{rgb}{0.0, 0.5, 0.0} % 自定义颜色
\usepackage{arydshln}
\usepackage{tikz}
\definecolor{table_yellow}{RGB}{254,243,222}
\definecolor{table_blue}{RGB}{237,245,255}
\definecolor{table_green}{RGB}{234,245,234}
\usepackage{multirow}
\usepackage{wrapfig}
\usepackage{booktabs}
\usepackage{subcaption}
\usepackage{booktabs, graphicx, caption}
\usepackage{tablefootnote}

\def\paperID{}
\def\confName{CVPR}
\def\confYear{2026}

\title{Mimic Human Cognition, Master Multi-Image Reasoning: A Meta-Action Framework for Enhanced Visual Understanding}

\author{
Jianghao Yin$^{1\dagger}$, Qingbin Li$^{2}$, Kun Sun$^{2}$, Cheng Ding$^{1}$, Jie Wang$^{2}$,
Qin Chen$^{1*}$, Jie Zhou$^{1}$,\\ Nan Wang$^{2}$, Changqing Li$^{2}$,
Pei Wu$^{2}$, Jian Xu$^{2*}$, Zheming Yang$^{2}$, Liang He$^{1}$\\
$^{1}$East China Normal University \quad $^{2}$ByteDance\\
{\tt\small jhyin@stu.ecnu.edu.cn, qchen@cs.ecnu.edu.cn, wukang.0325@bytedance.com}\\
\\
\footnotetext[1]{Work done during internship at ByteDance}
\footnotetext[2]{Corresponding Authors}
}

\begin{document}
\maketitle
\input{sec/0_abstract}

\begingroup
\renewcommand\thefootnote{*}
\footnotetext{Corresponding Authors.}
\renewcommand\thefootnote{$\dagger$}
\footnotetext{Work done during internship at ByteDance.}
\endgroup

\input{sec/1_intro}
\input{sec/2_related}

\input{sec/3_method}
\input{sec/4_expsetting}
\input{sec/5_exp}

\input{sec/6_conclusion}
\input{sec/7_Acknowledgment}

{
    \small
    \bibliographystyle{ieeenat_fullname}
    \bibliography{main}
}

% WARNING: do not forget to delete the supplementary pages from your submission 
\input{sec/X_suppl}

\end{document}

%% file: sec/0_abstract.tex
\begin{abstract}
While Multimodal Large Language Models (MLLMs) excel at single-image understanding, they exhibit significantly degraded performance in multi-image reasoning scenarios. Multi-image reasoning presents fundamental challenges including complex inter-relationships between images and scattered critical information across image sets. Inspired by human cognitive processes, we propose a Cognition-Inspired Meta-Action Framework (CINEMA), which decomposes multi-image reasoning into five structured meta-actions: Global, Focus, Hint, Think, and Answer, explicitly modeling the sequential cognitive steps humans naturally employ.
For cold-start training, we introduce a Retrieval-Based Tree Sampling strategy that generates high-quality meta-action trajectories to bootstrap the model with reasoning patterns. 
During reinforcement learning, we adopt a two-stage paradigm: an exploration phase with Diversity-Preserving Strategy to avoid entropy collapse, followed by an annealed exploitation phase with DAPO to gradually strengthen exploitation.
To train our model, we construct a dataset of 56k cold-start and 58k reinforcement learning instances spanning multi-image, multi-frame, and single-image tasks. We conduct extensive evaluations on multi-image reasoning benchmarks, video understanding benchmarks, and single-image benchmarks, achieving competitive state-of-the-art performance on several key benchmarks. 
Our model surpasses GPT-4o on the MUIR and MVMath benchmarks and notably outperforms specialized video reasoning models on video understanding benchmarks, demonstrating the effectiveness and generalizability of our human cognition-inspired reasoning framework.
\end{abstract}

%% file: sec/1_intro.tex
\section{Introduction}
\label{sec:intro}

Multimodal Large Language Models (MLLMs) have demonstrated remarkable capabilities in single-image understanding tasks \citep{qwen2.5vl,Internvl2.5,gpt4o,MMBenchmarksurvey,MLLMsurvey}, with extensive research focusing on enhancing models' single-image reasoning abilities \citep{multimodal_reasoning_survey,Vision-r1,VLLA,R1-onevision}. However, real-world applications often involve processing multiple images simultaneously, such as in e-commerce, autonomous driving, and video content understanding. Despite their success in single-image tasks, MLLMs exhibit degraded performance when handling multi-image reasoning scenarios \citep{MUIR,MMIU}.

Multi-image reasoning presents two fundamental challenges. First, images often exhibit complex inter-relationships: semantic associations, spatial arrangements, temporal sequences, that are crucial for task completion yet require sophisticated integration beyond isolated image processing \citep{VISC,MMIU}. Second, critical information may be scattered across specific images within larger sets, demanding precise identification and focus on relevant visual content while filtering out distractors.

Human cognition offers valuable insights for these challenges. When approaching complex multi-image reasoning tasks, humans employ a systematic process grounded in established cognitive principles. Research shows that humans effectively integrate both global structures and local details during visual perception \cite{cog1,cog2}, enabling comprehensive problem understanding across multiple levels. Additionally, studies demonstrate that explicitly identifying and articulating key information significantly enhances performance on complex reasoning tasks \cite{cog3}. This natural cognitive process suggests that artificial reasoning systems would benefit from cognitive frameworks that explicitly model these human-like reasoning patterns.

Motivated by these observations, we propose a
\textbf{C}ognition-\textbf{In}spir\textbf{e}d \textbf{M}eta-\textbf{A}ction Framework (CINEMA) that addresses multi-image reasoning through three key innovations. First, we introduce a set of five meta actions: Global, Focus, Hint, Think, and Answer, which systematically guide models through human-inspired reasoning processes. These meta actions provide a structured cognitive framework that enables models to effectively navigate the complexities of multi-image reasoning by explicitly modeling the sequential cognitive steps that humans naturally employ.
Second, we develop a Retrieval-Based Tree Sampling strategy that mirrors human learning dynamics through a student-teacher paradigm. This approach generates diverse, high-quality reasoning trajectories by first allowing a student model to attempt initial solutions, then having a teacher model refine these attempts, and finally retrieving alternative solution paths from a database of reasoning trajectories.
This process not only ensures the quality of training data but also refines the reasoning trajectories, enabling the model to generate reasoning patterns that more closely resemble human-like thinking.
Third, we design a novel two-stage reinforcement learning approach to optimize the reasoning process while maintaining trajectory diversity. We observe that standard reinforcement learning often suffers from entropy collapse \citep{wang2025beyond,cui2025entropy,CURE}, where policies become overly deterministic and lose exploration capacity over time. To address this challenge, our first stage employs \textbf{D}iversity-\textbf{P}reserving \textbf{S}trategy (DPS) with a trajectory homogeneity penalty to maintain sufficient exploration and prevent premature convergence to suboptimal solutions. The second stage then applies decoupled
clip and dynamic sampling policy optimization (DAPO) \citep{yu2025dapo} to gradually transition toward more focused behaviors, effectively balancing the exploration-exploitation trade-off throughout the training.

To train our model, we construct a high-quality training dataset comprising 56k cold-start instances and 58k reinforcement learning instances. Each cold-start instance contains two distinct reasoning trajectories to provide diverse supervision signals during initial training. The dataset encompasses three categories of visual reasoning tasks: multi-image tasks, multi-frame tasks and single-image tasks. Our main contributions are as follows:

\begin{itemize}[leftmargin=*, align=left]

\item We propose a human cognition-inspired reasoning framework that decomposes complex multi-image reasoning into five structured meta actions (Global, Focus, Hint, Think, Answer). This framework systematically models the sequential cognitive processes that humans naturally employ when solving multi-image reasoning tasks, providing explicit guidance for models to navigate complex visual reasoning scenarios.

\item We introduce a novel Retrieval-Based Tree Sampling strategy that generates diverse, high-quality trajectories through student-teacher interactions, coupled with a two-stage reinforcement learning paradigm: Diversity-Preserving Strategy with trajectory homogeneity penalty to maintain exploration, followed by DAPO to consolidate performance while preserving learned diversity.

\item We construct a comprehensive training dataset with 56k cold-start instances where each contains two reasoning trajectories, and 58k reinforcement learning instances across multi-image, multi-frame, and single-image tasks.

\item We conduct extensive evaluations across multiple benchmarks spanning multi-image reasoning, video understanding, and single-image tasks. Our method achieves state-of-the-art performance on various benchmarks and notably outperforms specialized video reasoning models on video understanding tasks, demonstrating the effectiveness and generalizability of our approach.
\end{itemize}

\begin{figure*}[]
  \centering
  \includegraphics[width=1\textwidth]{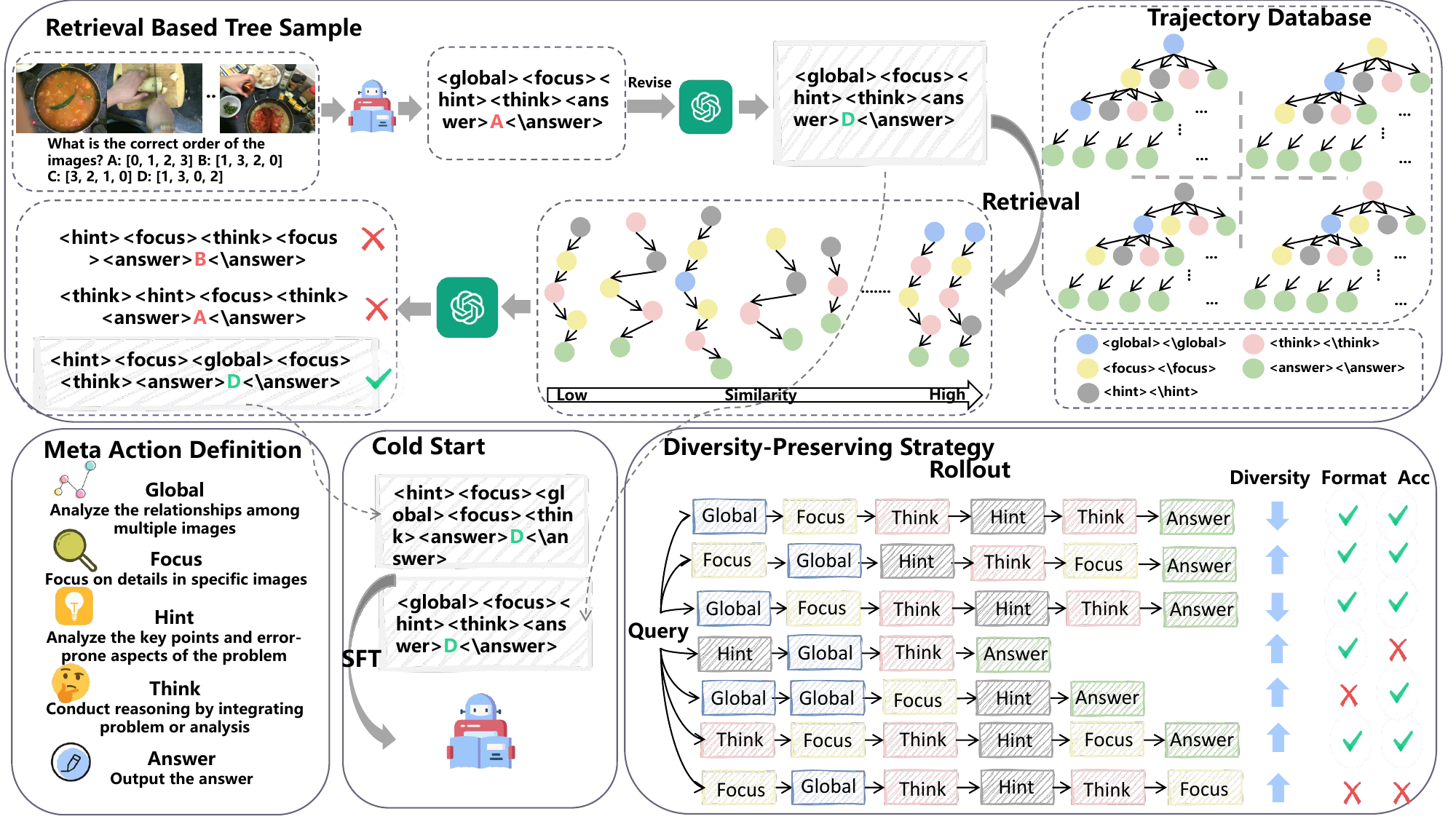}\\
  \caption{Overview of CINEMA.}
  \label{framework}
\end{figure*}

%% file: sec/2_related.tex
\section{Related Work}

\paragraph{Multimodal Reasoning.}

MLLMs typically consist of a vision encoder \cite{clip, fuzzyvit, li2025unleashing}, a projection layer, and a large language model, and are widely utilized in scenarios such as video understanding \cite{Videor1}, digital humans \cite{omnihuman, showmaker}, and medical image analysis \cite{medicalmllm}.
Recent works have enhanced MLLM reasoning capabilities \citep{Vision-r1,insight,sketchpad,vltools,look_back}, but most focus on single-image scenarios. VISC \citep{VISC} propose a Focus-Centric Visual Chain that decomposes multi-image tasks into sequential sub-questions.  MIA-DPO \citep{MIA-DPO} augments single-image datasets with unrelated images for preference optimization, but primarily handles cases where questions involve only single images within multi-image contexts.
Authentic multi-image reasoning requires models to analyze individual images while comprehending holistic relationships among all images. Inspired by human cognition, we propose a framework that effectively navigates both local image analysis and global inter-image relationships.

% \subsection{Reinforcement Learning for Reasoning and Entropy Control.} 
\paragraph{Reinforcement Learning.} Early reinforcement learning approaches for foundation models relied on Reinforcement Learning from Human Feedback (RLHF), which required training a separate reward model and human-labeled preference data \citep{ouyang2022training,hunter2004mm}. Direct Preference Optimization (DPO) \citep{rafailov2023direct} simplified this pipeline but still depended on preference annotations. More recently, large-scale pure RL methods have shown strong gains in reasoning, with outcome-based rewards alone proving effective \citep{guo2025deepseek,team2025kimi,hu2025open,liu2025understanding,yan2025learning,chen2025bridging}. To regulate exploration, many approaches add entropy or KL regularization \citep{he2025skywork,liu2025prorl}, introduce entropy bonuses through reward shaping \citep{cheng2025reasoning}, or apply stabilizing heuristics such as loss reweighting \citep{wang2025beyond,cui2025entropy} and clip-higher mechanisms \citep{yu2025dapo}. While these methods focus on entropy within a single response, others encourage diversity across responses, e.g., via embedding-based distance measures \citep{chen2025dra} or enforcing dissimilarity in generated answers \citep{chen2025seed}. Our approach builds on this line of work but emphasizes diversity at the meta-action level for entropy control.

%% file: sec/3_method.tex
\section{Method}

The famework of our method is shown in Figure \ref{framework}. We first define five structured meta actions: Global, Focus, Hint, Think, and Answer, which model human cognitive processes (Section \ref{Sec: action definition}). We then propose Retrieval-Based Tree Sampling to generate diverse, high-quality training trajectories via student-teacher interactions (Section \ref{sec3.2}), and construct a comprehensive dataset with 56k cold-start and 58k reinforcement learning instances (Section \ref{dataset}). Finally, we introduce a two-stage training paradigm: in the first stage, Diversity-Preserving Strategy prevents entropy collapse and maintains trajectory diversity during reinforcement learning, and in the second stage, DAPO anneals the policy toward exploitation to consolidate performance (Section \ref{DIPO}).

\subsection{Meta Action Definition}
\label{Sec: action definition}

\textbf{Global.} This meta action simulates how humans typically approach complex problems by first reading through the entire question to grasp its overall structure. When dealing with multi-image input tasks, there may be temporal, spatial, semantic, or other relationships between the images. This action helps the model identify and leverage these inter-image dependencies to enhance understanding and reasoning.

\noindent \textbf{Focus.} This meta action simulates how humans tackle complex problems by concentrating on analyzing key information relevant to the question. In the context of multi-image reasoning, critical clues may reside in a specific image. The model should therefore focus its analysis on that image and pay close attention to salient visual details.

\noindent \textbf{Hint.} This meta action simulates how humans improve accuracy by summarizing key points and error-prone aspects of a problem when solving tasks. In multi-image reasoning tasks, similarly, there often exists misleading or easily confusable information between images. 

\noindent \textbf{Think.} This meta action simulates how humans engage in internal reasoning by actively processing acquired information to formulate solutions or hypotheses. It involves analyzing the relationships between provided clues, leveraging prior knowledge, and performing logical inference. 

\noindent \textbf{Answer.} This meta action outputs the final answer based on all prior analytical insights and reasoning outcomes. It is the final action in the compliant trajectory.

\subsection{Retrieval Based Tree Sample}
\label{sec3.2}
To effectively leverage the defined meta actions for multi-image reasoning enhancement, we propose a novel cold-start data sampling strategy called Retrieval-Based Tree Sampling. This approach is inspired by human learning mechanisms, where students first attempt problems independently before receiving guidance from teachers who first refine their initial approach and then introduce alternative solution pathways.

With the meta actions defined in Section \ref{Sec: action definition}, we maintain several meta action trees, each containing multiple reasoning trajectories. 
The root nodes of meta action trees can be \texttt{Global}, \texttt{Focus}, \texttt{Hint}, or \texttt{Think}, indicating possible starting actions for reasoning. 
Every trajectory in these trees terminates with the \texttt{Answer} meta action, forming complete reasoning paths from problem comprehension to solution derivation. These trees serve as a database of diverse reasoning strategies that can be retrieved and adapted for new problems.  The Retrieval-Based Tree Sampling process is shown as follows: 

\textbf{Step 1. Initial Trajectory Generation.}
We first prompt a smaller model (student model) to perform initial reasoning on the given task using meta actions. This generates an initial trajectory regardless of whether the final answer is correct or incorrect. This step mirrors how students first attempt to solve problems using their existing knowledge and reasoning patterns.

\textbf{Step 2. Teacher-Guided Trajectory Refinement.}
The initial trajectory from Step 1 is provided to a stronger model (GPT-4o, serving as the teacher model). The teacher model follows the student's action trajectories and reason again, similar to how human teachers guide students by first understanding their thought processes and then providing corrections. This produces a correct trajectory that maintains the original action trajectories while ensuring accuracy.

\textbf{Step 3. Retrieval-Based Diverse Sampling.}
To enrich the learning experience and expand the exploration space for subsequent reinforcement learning, we perform retrieval-based sampling from our trajectory tree database. Starting from trajectories with low similarity to the initial one from Step 2, we progressively search through increasingly similar trajectories until identifying an alternative correct reasoning path. This process ensures that each training instance is associated with two distinct correct trajectories. Similarity is measured by encoding trajectories via BGE \cite{BGE} and then calculating their cosine similarity.

\subsection{Dataset Construction}
\label{dataset}
To train our model, we construct a high-quality training dataset that supports both cold-start initialization and reinforcement learning phases. Our dataset encompasses three primary categories: multi-image tasks in which the number of input images is at least two, multi-frame tasks that involve reasoning over sequential frames from videos or time-series visual data, and single-image tasks that the number of input image is only one. 
All the data is obtained through existing open-source multi-modal datasets. More details about dataset is shown in Appendix.

The key distinction between our cold-start and reinforcement learning dataset splits lies in the trajectory generation process described in Section \ref{sec3.2}. Cold-start training data consists of problems where GPT-4o successfully provides correct answers during Step 2, and for these instances, we proceed to Step 3 (retrieval-based diverse sampling) to obtain two distinct correct reasoning trajectories per problem that serve as supervised learning targets for cold start training. In contrast, reinforcement learning data comprises problems where GPT-4o fails to produce correct answers during Step 2, and these challenging cases are reserved for reinforcement learning.

\subsection{Balancing Exploration and Exploitation via Two-Stage Optimization}
\label{DIPO}

A critical challenge in reinforcement learning for reasoning is policy entropy collapse which limits exploration and generalization capacity. We address this through a two-stage training paradigm: first maintaining trajectory diversity to preserve exploration, then gradually shifting toward exploitation to consolidate performance.

\textbf{Diversity-Preserving Strategy.} In the first stage,
we aim to prevent entropy collapse by maintaining diversity at the meta-action level. To this end, we propose Diversity-Preserving Strategy (DPS) which is built on DAPO \citep{yu2025dapo} (more details about DAPO is shown in Appendix). Our central hypothesis is that encouraging a variety of solution strategies can better leverage the model’s potential and improve its generalization performance like human.

We operationalize this by promoting diverse responses for questions that the model answers correctly. 
To this end, we define the reward as a weighted combination of accuracy and format validity:
\begin{equation}
R = 0.5 \cdot \left[ R_\text{acc} \cdot \left(R_\text{acc} - \frac{N-1}{G-1} \cdot 0.1 \right) \right] 
    + 0.5 \cdot R_\text{format},
\end{equation}

where $R_\text{acc}$ and $R_\text{format}$ are binary indicators:
\begin{equation}
\begin{gathered}
R_\text{acc} =
\begin{cases}
1, & \text{if the answer is correct}, \\
0, & \text{otherwise},
\end{cases}\\
R_\text{format} =
\begin{cases}
1, & \text{if all meta actions in the response are valid}, \\
0, & \text{otherwise}.
\end{cases}
\end{gathered}
\end{equation}

\noindent Here, $G$ denotes the group size used in sampling, and $N$ represents the number of 
trajectories within the group that share identical meta-action patterns. Intuitively, 
the penalty term $\tfrac{N-1}{G-1}$ discourages over-reliance on homogeneous trajectories, 
thereby encouraging the model to maintain diversity across solutions.
This design ensures that correct answers are not only accurate but also exhibit a broad spectrum of 
solution strategies, thereby enhancing the model’s generalization. In practice, to perform dynamic 
sampling as in DAPO, we use $R_\text{acc}$ as the filtering criterion rather than the combined reward $R$.

\textbf{Annealed Exploitation.} In the second stage, we employ DAPO with an annealing schedule to gradually shift from exploration to exploitation, leveraging the diversity obtained in stage one while consolidating performance gains. This two-stage approach maintains higher entropy levels throughout training compared to standard methods, as validated by our Pass@K experiments.

%% file: sec/4_expsetting.tex
\section{Experiment Setup}

\subsection{Benchmark and Baselines}

\textbf{Benchmark. }To ensure a comprehensive evaluation, we examine the performance of our method across a broad spectrum of benchmarks, encompassing both multi-image and single-image types. Specifically, for multi-image evaluations, we cover \textbf{multi-image comprehensive benchmarks} (including MUIR \citep{MUIR}, MMIU \citep{MMIU}, and Mantis-Eval \citep{Mantis}), \textbf{multi-image reasoning benchmarks} (including MV-MATH \citep{mvmath}, MIRB \citep{MIRB} and EMMA \citep{EMMA}), \textbf{video comprehensive benchmarks} (including MVBench \citep{MVBench}, and VideoMME \citep{VideoMME}) and \textbf{video reasoning benchmarks} (including VideoMMMU \citep{VideoMMMU}). For single-image evaluations, we include \textbf{single-image comprehensive benchmarks} (including MMMU-Pro \citep{MMMU-Pro} and M3COT \citep{m3cot}) as well as \textbf{mathematics reasoning benchmarks} (including MM-Math \citep{MM-MATH}, Math-Vision \citep{Math-Vsion}, and MathVista \citep{Mathvista}). Accuracy is reported as the metric for all these benchmarks.

\textbf{Baslines. }We compare our method’s against four categories of models: closed-source MLLMs, including GPT-4V \citep{gpt4v}, Gemini-1.5-Pro \citep{gemini}, and GPT-4o \citep{gpt4o}; open-source general MLLMs, including OpenFlamingo-v2 \citep{awadalla2023openflamingo}, LLaVAv1.6 \citep{liu2024llavanext}, VILA1.5 \citep{vila}, LLaVA-OneVision \citep{LLaVA-OneVision}, InternVL2.5 \citep{Internvl2.5}, and Qwen2.5-VL \citep{qwen2.5vl}; and multi-image/video enhanced models, including Mantis-Idefics \citep{Mantis}, mPLUG-Owl3 \citep{mPLUG-Owl3}, LLaVA-NeXT-Interleave \citep{Llava-next-interleave}, CMMCOT \citep{cmmcot}, MIA-DPO \citep{MIA-DPO}, VISC \citep{VISC}, VideoR1 \citep{Videor1}, VideoRFT \citep{Videorft} and TW-GRPO \citep{twgrpo}; Single-image reasoning models: Mulberry  \citep{mulberry}, R1-Onevision \cite{R1-onevision}, VLAA-Thinker \citep{VLLA}, VisonR1 \citep{Vision-r1}, MixedR1  \citep{mixedr1}.

\subsection{Implementation Details}
We select Qwen2.5VL 7B as our backbone model. During the cold-start training phase, the model is trained for two epochs with a learning rate of $1\times10^{-5}$. 
We employ a two-stage reinforcement learning procedure. The first stage consists of 700 steps of DPS-based entropy enhancement, followed by 300 steps of DAPO-based annealed exploitation. In the subsequent reinforcement learning stage, both the KL-divergence and entropy regularization terms are omitted. Rollouts are generated using a batch size of $64$, a temperature of $1.0$, and $8$ rollouts per prompt. For policy optimization, an update batch size of $32$ is adopted.
Regarding reward design, we incorporate domain-specific validation mechanisms: 
\texttt{math\_verify} \citep{Kydlicek_Math-Verify_Math_Verification} and \texttt{mathruler} \citep{mathruler} are employed to evaluate answers in mathematical problem-solving, 
whereas exact string matching is applied to non-mathematical tasks. 
To ensure structural consistency, format rewards are introduced by imposing constraints on the response space, 
requiring outputs to conform to a valid meta-action trajectory. 
Specifically, for single-image inputs, the \texttt{Global} action is disallowed.
During inference, we set the decoding hyperparameters as follows: temperature $=0.6$, top-$p=0.7$, 
and a maximum of $1024$ generated tokens. 
More implementation details are in the Appendix.

%% file: sec/5_exp.tex
\section{Experiments}

\subsection{Results on Multi-Image Benchmark}

Table \ref{multi-imgae} presents the experimental results on multi-image benchmarks, where our model demonstrates significant improvements over Qwen2.5VL across all benchmarks, achieving state-of-the-art performance on MUIR, MVMath, EMMA, VideoMME, and VideoMMMU benchmarks. Remarkably, our model surpasses the closed-source GPT-4o on both MUIR and MVMath benchmarks. On  multi-image comprehensive benchmarks, our model achieves 13.7\% improvement over Qwen2.5VL on the MUIR benchmark and 6.9\% improvement on MIRB. These multi-image benchmarks encompass diverse multi-image tasks, demonstrating our model's robust capability in processing multi-image inputs. On multi-image resoning benchmarks, MVMath is a mathematics dataset with multi-image inputs, while EMMA encompasses multiple academic disciplines. These benchmarks require strong reasoning capabilities from the model. Our model achieves 10.2\% improvement on MVMath and 8.9\% improvement on EMMA, reflecting enhanced reasoning capabilities attributed to CINEMA, which simulates human-like reasoning processes through structured meta-action trajectory and cross-image relationship modeling. Notably, our model surpasses specialized video reasoning models across all three video benchmarks. This demonstrates our model's superior performance in handling temporal multi-image data, suggesting that our approach effectively captures both spatial and temporal dependencies inherent in sequential visual information.

\begin{table*}[htbp]
\setlength{\tabcolsep}{1pt}
\small
% \scriptsize
\centering

\begin{tabular}{ccccccccccc}
\toprule
\textbf{Model} & \textbf{MUIR} & \textbf{MMIU} & \textbf{MVMATH} & \textbf{EMMA} & \textbf{MIRB} & \textbf{Mantis} & \textbf{MVBench} & \textbf{VideoMME} & \textbf{VideoMMMU}  & \textbf{Overall} \\
\hline
\multicolumn{11}{c}{\textit{Closed-Source MLLMs}} \\
\hline
GPT-4V          & - & - & 24.5 & - & - & 62.7 & 43.5  & 59.9 &- &- \\
Gemini-1.5-Pro  & - & - & 29.1 & - & - & - & - & 71.9 &53.9 & -\\
GPT-4o          & 68.0 & 55.7  & 32.1 & 32.7 & - & - & - & 75.0 &61.2 &- \\
\hline
\multicolumn{11}{c}{\textit{Open-Source General MLLMs}} \\
\hline
OpenFlamingo-v2 9B & 22.3 & 23.7 & - & - & 28.8 & 12.4 & 7.9 &- &- &-\\
LLaVA 1.6 7B       & 27.4 & 22.2 & - & - & 29.8 & 45.6 & 40.9 &- &- &-\\
VILA1.5 8B         &33.1 &32.5 &- &- &36.5 & 51.2 &49.4 &20.9 &-  &-\\
LLaVA-OneVision 7B &41.8 &40.3 &19.1 &- & 51.2 &  64.2 &56.7 &- &- &-\\
InternVL2.5 8B     & 51.1 & 46.7  & 18.8 & 21.0 & 52.5 & 67.7  & \textbf{72.0} &56.1 &35.2 &46.8\\
Qwen2.5-VL-7B      & 57.9 & 50.6 & 26.7 & 20.4 & 48.3 & 64.5 & 62.6 &56.7  &45.8 &48.2\\
\hline
\multicolumn{11}{c}{\textit{Multi-image/Video Enhancing MLLMs}} \\
\hline
Mantis-Idefics2 8B & 44.5 & 45.6 & 15.5 & 20.3 & 34.8 & 57.1 &  51.4 &42.6 &19.3 &36.8 \\
LLaVA-NeXT-Interleave 7B & 31.1  &47.3 &14.7 &19.0 &39.3 &62.7 & 53.1 &47.2 &23.2 &37.5\\
mPLUG-Owl3 8B & 34.0 & 39.7 & 18.7 & 24.8 & 41.2 & 63.1 & 54.5 &53.5 &32.0 &40.2 \\
MIA-DPO 7B & - & - & - & - & - & 60.4 & 63.6 &- &- &-\\
CcDPO 7B  & 44.8 & - & - & - & \textbf{60.7} & 69.1 &  - &- &- &- \\
VISC 7B   & 44.5 & 52.8 & - & - &60.2 & 69.1 & 68.0 & -& - &-\\
VideoR1 7B & - & - & - & - & - & - & 63.6 & 57.4 &49.8 &- \\ 
VideoRFT 7B& 56.6 & 44.5 & 25.1 &17.8 &46.7 &56.7 &62.1 &59.8 &51.1 &46.7\\
TW-GRPO 7B &55.9  &44.9  &28.2  &22.5 &24.3 &49.8 &63.3 &55.1 &40.8 &42.8\\
\hline
Ours & \textbf{71.6} & \textbf{53.3} & \textbf{36.9} & \textbf{29.3} & 55.2 & 67.7 &66.5 &59.4 &49.0 &\textbf{54.3} \\
$\Delta$ (vs Qwen2.5VL 7B) & \textcolor{red}{+13.7} & \textcolor{red}{+2.7} & \textcolor{red}{+10.2} & \textcolor{red}{+8.9} & \textcolor{red}{+6.9} & \textcolor{red}{+3.2} & \textcolor{red}{+3.9} & \textcolor{red}{+2.7} & \textcolor{red}{+3.2} & \textcolor{red}{+6.1}\\
\hline
Ours [with DPS] &67.9 & 52.2 & 35.1 &28.4 &54.4  &\textbf{71.0} &67.1 &60.2 &\textbf{51.6} &54.2 \\
$\Delta$ (vs Qwen2.5VL 7B) & \textcolor{red}{+10.0} & \textcolor{red}{+1.6} & \textcolor{red}{+8.4} & \textcolor{red}{+8.0} & \textcolor{red}{+6.1} & \textcolor{red}{+6.5} & \textcolor{red}{+4.5} & \textcolor{red}{+3.5} & \textcolor{red}{+5.8} & \textcolor{red}{+6.0}\\
\hline
Ours [with DPS and annealing] & 71.0 & 52.2 & 35.0 & 28.6 & 55.7 & 68.4 &66.8 &\textbf{61.0} &50.1 &\textbf{54.3} \\
$\Delta$ (vs Qwen2.5VL 7B) & \textcolor{red}{+13.1} & \textcolor{red}{+1.6} & \textcolor{red}{+8.3} & \textcolor{red}{+8.2} & \textcolor{red}{+7.4} & \textcolor{red}{+3.9} & \textcolor{red}{+4.2} & \textcolor{red}{+4.3} & \textcolor{red}{+4.3} & \textcolor{red}{+6.1}\\
\bottomrule
\end{tabular}
\caption{Performance on multi-image/video benchmark. Ours indicates training with DAPO, Ours [with DPS] indicates training with DPS, and Ours [with DPS and annealing] indicates training with two-stage RL, where all models are trained for the same steps.}

\label{multi-imgae}

\end{table*}

\subsection{Results on Single-Image Benchmark}

Table \ref{single-imgae} presents the results on single-image benchmarks, where our model demonstrates equally strong capabilities. Our model achieves superior overall performance compared to existing models specifically designed for single-image reasoning. This validates the generalizability of our approach, proving its effectiveness not only for multi-image scenarios but also for single-image tasks. On comprehensive single-image benchmarks, our model achieves 3\% improvement on MMMU-Pro and 3.8\% improvement on M3COT, surpassing the closed-source models GPT-4V and GPT-4o on M3COT. On math benchmarks, our model attains state-of-the-art performance on MM-Math and achieves comparable performance to existing models specialized in single-image reasoning across other mathematical benchmarks.

\begin{table*}[htbp]
\setlength{\tabcolsep}{1pt}
\small
\centering
\begin{tabular}{ccccccccccc}
\toprule
\textbf{Model} & \textbf{MMMU-Pro} & \textbf{M3COT} &\textbf{MM-IQ} & \textbf{MM-Math} & \textbf{Math-Vision} & \textbf{MathVista} & \textbf{MathVerse} &  \textbf{Overall} \\
\hline
\multicolumn{11}{c}{\textit{Closed-Source MLLMs}} \\
\hline
GPT-4V          & - & 62.60 & - &  23.1 &22.76 & 49.9 & 39.4  & - \\
Gemini-1.5-Pro  & 51.47 & - &  26.86 & - & - & - & - & - \\
GPT-4o          & 56.13 & 55.7  & 26.87 & 31.8 & - & - & - & -\\
\hline
\multicolumn{11}{c}{\textit{Open-Source General MLLMs}} \\
\hline
LLaVA-OneVision 7B  & - & - & - & - & - & 63.2 & 26.2 & - \\
InternVL2.5 8B     & 34.3 & - & - & - & 22.0 & 64.4 & 39.5 & - \\
Qwen2.5VL 7B      & 38.0 & 60.1 & 26.1 & 36.4 & 19.5 & 65.3 & 40.4 & 40.8 \\
\hline
\multicolumn{11}{c}{\textit{Reasoning MLLMs}} \\
\hline
Mulberry 7B & - & - & - & 23.7 & - & 63.1 & - &- \\

R1-Onevision 7B & 33.9 & 57.3 & 25.1 & 32.9 & 29.9 & 64.1 & 46.4 &41.4 \\
VLAA-Thinker 7B & 39.5 & 61.3 & 26.3 & 39.0 & 26.4 & 68.0 & 47.8 &44.0 \\
VisonR1 7B & 30.3 & 53.2 & 24.3 & 40.0 & 29.9 & \textbf{73.5} & \textbf{52.4} &43.4 \\
MixedR1 7B & 38.0 & 59.9 & 25.9 & 35.8 & \textbf{30.3} & 70.6 &40.8 & 43.0\\
\hline
Ours & 40.6 & 63.5 & 25.6 & \textbf{43.8} & 26.7 & 68.7 &49.4 &45.5 \\
$\Delta$ (vs Qwen2.5VL 7B) & \textcolor{red}{+2.6} & \textcolor{red}{+3.4} & \textcolor{red}{+0.5} & \textcolor{red}{+7.4} & \textcolor{red}{+7.2} &\textcolor{red}{+3.4} & \textcolor{red}{+9.0} & \textcolor{red}{+4.7}\\
\hline
Ours [with DPS] &40.7 & \textbf{63.9} & 26.3 &43.4 &26.1 &70.0 &47.6 &45.4\\
$\Delta$ (vs Qwen2.5VL 7B) & \textcolor{red}{+2.7} & \textcolor{red}{+3.8} & \textcolor{red}{+0.2} & \textcolor{red}{+7.0} & \textcolor{red}{+6.6} &\textcolor{red}{+4.7} & \textcolor{red}{+7.2} & \textcolor{red}{+4.6}\\
\hline
Ours [with DPS and annealing] & \textbf{41.0} & 62.7 & \textbf{27.3} & 43.4 & 26.1 & 70.1 &48.5 &\textbf{45.6} \\
$\Delta$ (vs Qwen2.5VL 7B) &\textcolor{red}{+3.0} & \textcolor{red}{+2.6} & \textcolor{red}{+1.2} & \textcolor{red}{+7.0} & \textcolor{red}{+6.6} &\textcolor{red}{+4.8} & \textcolor{red}{+8.1} & \textcolor{red}{+4.8}\\

\bottomrule
\end{tabular}
\caption{Performance on single-image benchmark.Ours indicates training with DAPO, Ours [with DPS] indicates training with DPS, and Ours [with DPS and annealing] indicates training with two-stage RL.}
\label{single-imgae}
\end{table*}

\begin{figure}[]
    \centering
    \includegraphics[width=0.48\textwidth]{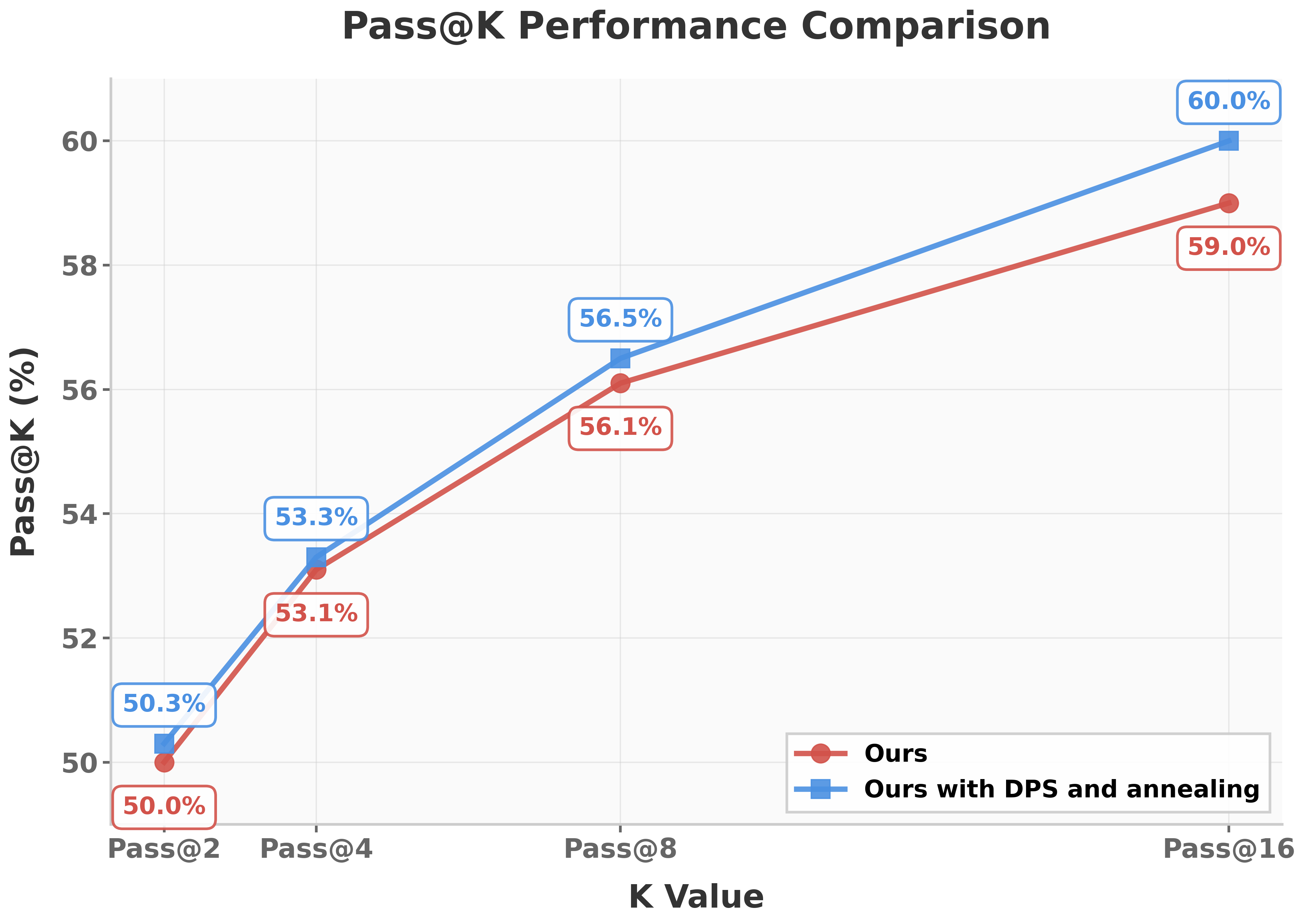}  
    \caption{Pass@K performance.}
    \label{fig:pass}
\end{figure}

\subsection{Results on Pass@K Setting}

To further validate the advantages of our proposed two-stage RL approach, we conduct Pass@K experiments on 7 multi-image and 7 single-image benchmarks, comparing models with and without DPS and annealing. We evaluate the accuracy across K inference attempts, where $K \in {2, 4, 8, 16}$, and a model is considered correct if at least one inference attempt produces the correct answer. We report the average accuracy in Figure \ref{fig:pass}. The results show that incorporating two-stage RL consistently outperforms the baseline across pass@2, pass@4, pass@8, and pass@16, further demonstrating the effectiveness of our two-stage RL method. After this training   paradigm, the model exhibits more diverse sampling behavior and achieves a higher performance ceiling.

\begin{table}[htbp]
    \centering
    % \small  
    \setlength{\tabcolsep}{2.5pt} 
    \renewcommand{\arraystretch}{1.0} 
    \begin{tabular}{l *{6}{c}}
        \toprule
        \textbf{Method} & \multicolumn{2}{c}{\textbf{MUIR}} & \multicolumn{2}{c}{\textbf{MMIU}} & \multicolumn{2}{c}{\textbf{EMMA}} \\
        \midrule
        Original & \multicolumn{2}{c}{57.9} & \multicolumn{2}{c}{50.6} & \multicolumn{2}{c}{20.4} \\
        Direct Prompting & \multicolumn{2}{c}{33.8} & \multicolumn{2}{c}{36.9} & \multicolumn{2}{c}{14.1} \\
        \midrule  
        & SFT & RL & SFT & RL & SFT & RL \\  
        \cmidrule(lr){2-3} \cmidrule(lr){4-5} \cmidrule(lr){6-7}  
        Conventional CoT & \textbf{59.0} & 70.0 & 49.9 & 51.6 & 21.2 & 26.9 \\
        Single Trajectory & 56.3 & 65.1 & 50.9 & 52.2 & 24.0 & 27.9 \\
        \textbf{Ours (Two Traj.)} & 58.2 & \textbf{71.6} & \textbf{51.9} & \textbf{53.3} & \textbf{24.8} & \textbf{29.3} \\
        \bottomrule
    \end{tabular}
    \caption{Ablation study on Retrieval-Based Tree Sampling.}
    \label{tab:ablation_study}
\end{table}

\subsection{Further Analysis}
{\setlength{\parskip}{2pt}
To conduct an in-depth analysis of our model, we propose 5 research questions and conduct detailed experiments:

\textbf{RQ1: Can diverse trajectories improve model performance?}
To validate the effectiveness of our proposed Retrieval-Based Tree Sampling strategy, which samples two different trajectories for each data point, we conduct comparative experiments on three benchmarks: MUIR, MMMU and EMMA. We compare against three baselines: (1) cold start training with only one trajectory then RL; (2) cold start training with conventional Chain-of-Thought (CoT) in the format of: \texttt{<think>reasoning here</think><answer>answer</answer>} then RL; and (3) directly prompting MLLMs to perform reasoning using meta actions without additional training. The experimental results are shown in Table \ref{tab:ablation_study}. The model trained with two trajectories achieves superior average performance compared to models trained with single trajectories and conventional CoT training. Moreover, the best results under RL are all achieved by the model trained with two trajectories. In comparison to the directly prompted model, we observe that the untrained model performs poorly in utilizing our defined meta actions, showing significant performance degradation relative to the original model. This demonstrates the necessity of constructing datasets for subsequent training.

\begin{figure*}[]
  \centering
  \begin{subfigure}[b]{0.32\linewidth}
    \includegraphics[width=\linewidth]{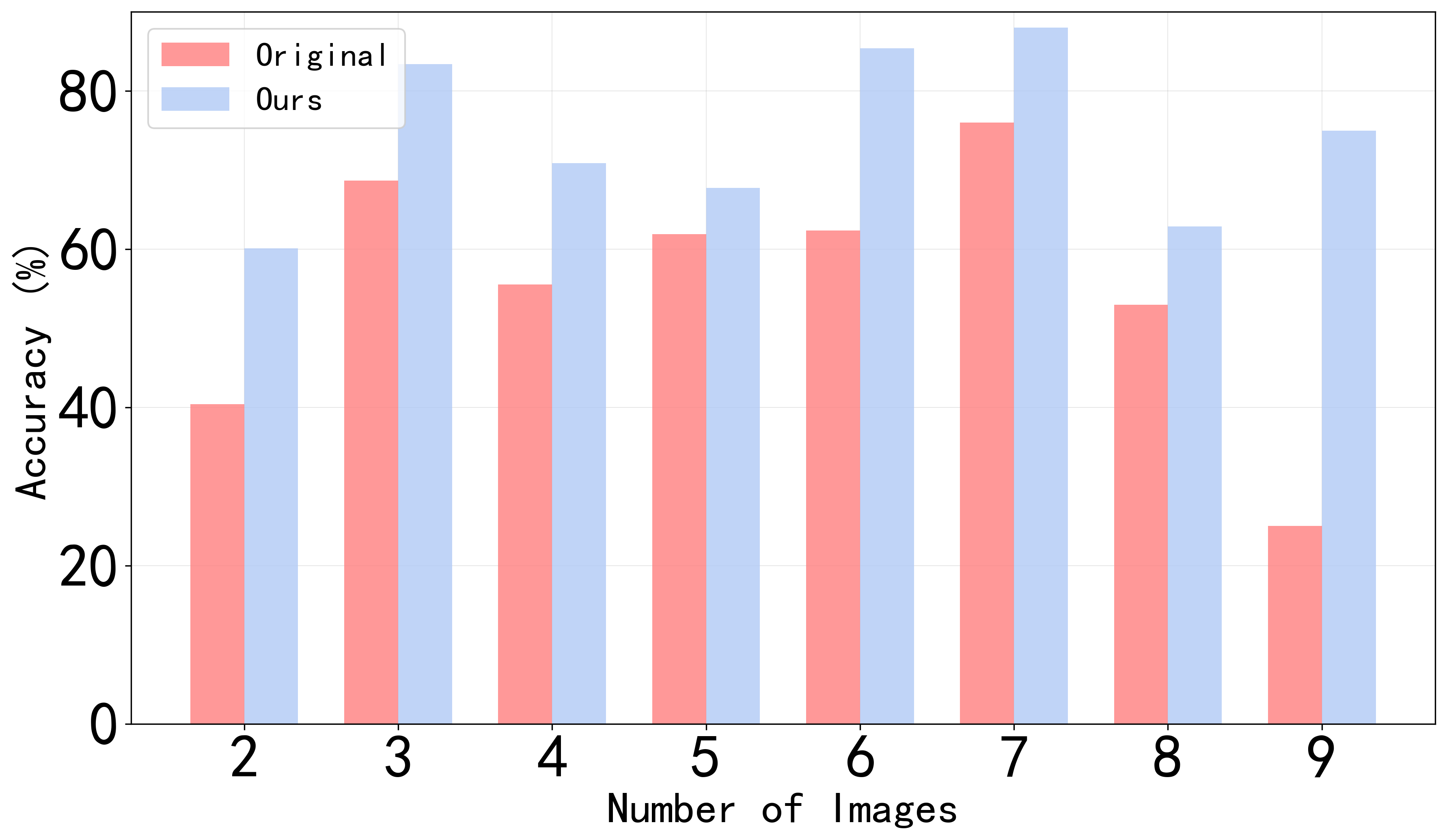}
    \caption{Results on MUIR.}
    \label{MUIR}
  \end{subfigure}
  \hfill
  \begin{subfigure}[b]{0.32\linewidth}
    \includegraphics[width=\linewidth]{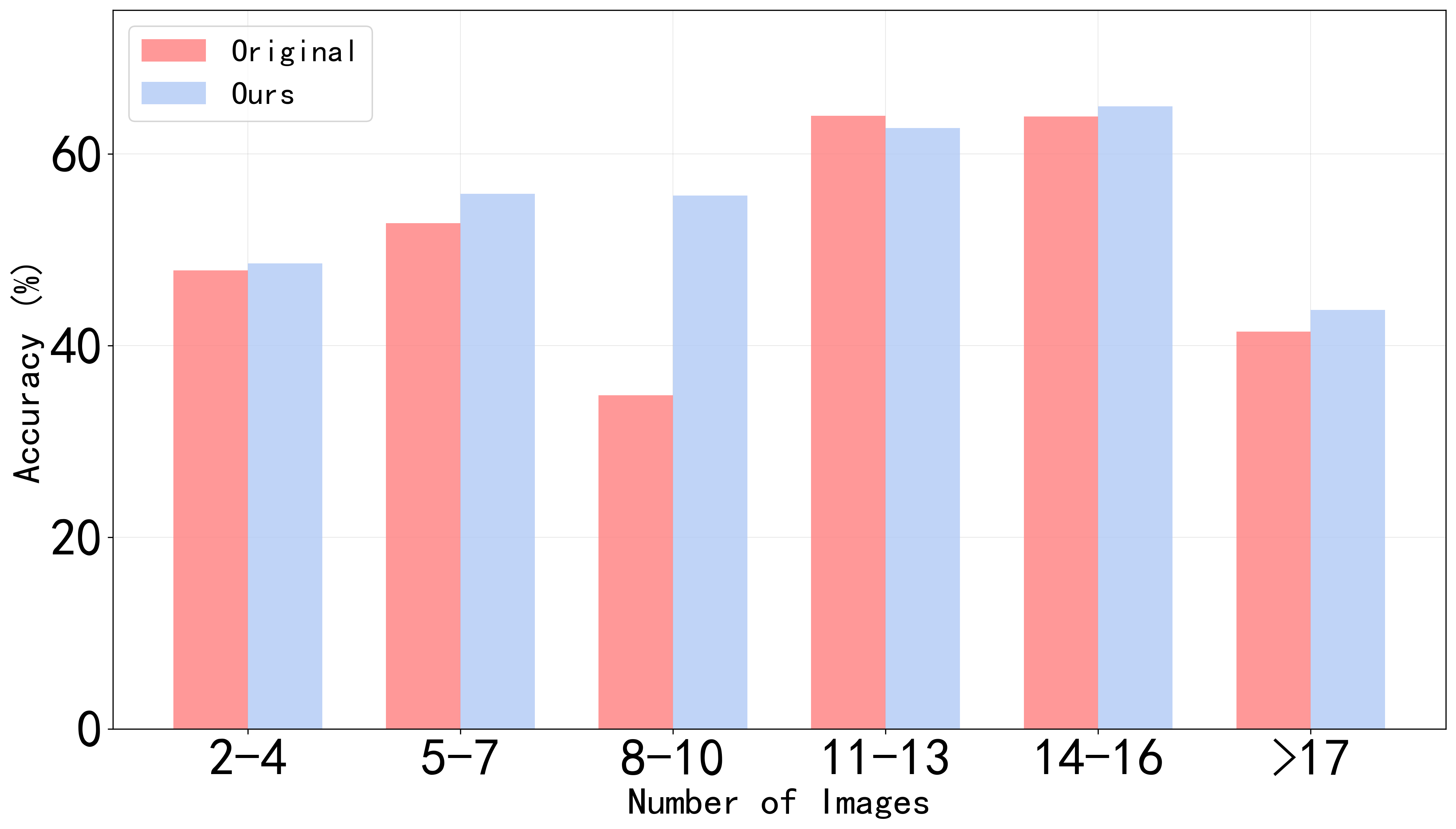}
    \caption{Results on MMIU.}
    \label{MMIU}
  \end{subfigure}
  \hfill
  \begin{subfigure}[b]{0.32\linewidth}
    \includegraphics[scale=0.15]{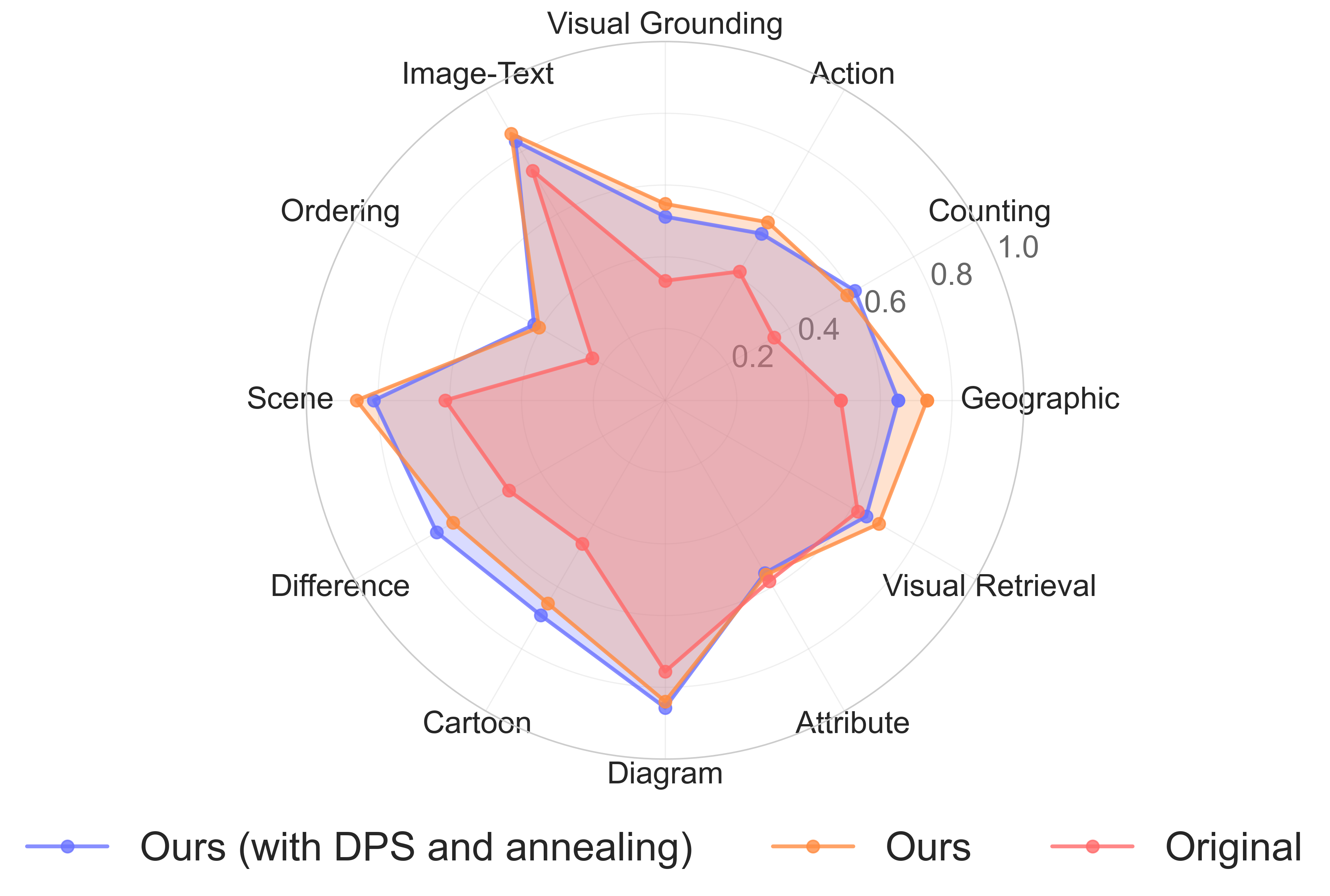}
    \caption{Results on different tasks.}
    \label{task}
  \end{subfigure}
  \caption{Results about RQ2 and RQ3.}
  \label{para}
\end{figure*}

\begin{figure*}[]
  \centering
  \begin{subfigure}[b]{0.32\linewidth}
    \includegraphics[width=\linewidth]{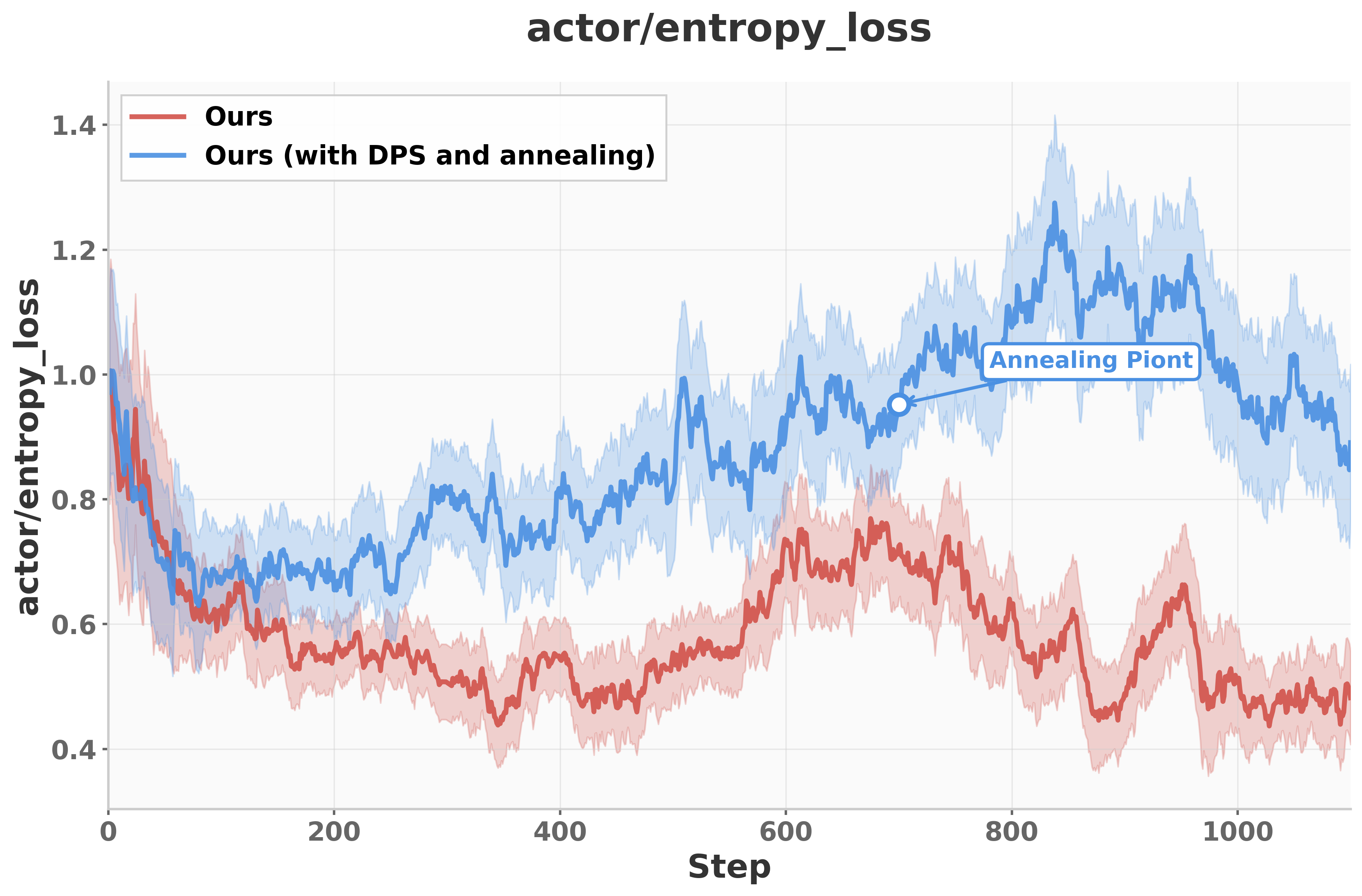}
    \caption{Entropy loss comparison.}
    \label{loss}
  \end{subfigure}
  \hfill
  \begin{subfigure}[b]{0.32\linewidth}
    \includegraphics[width=\linewidth]{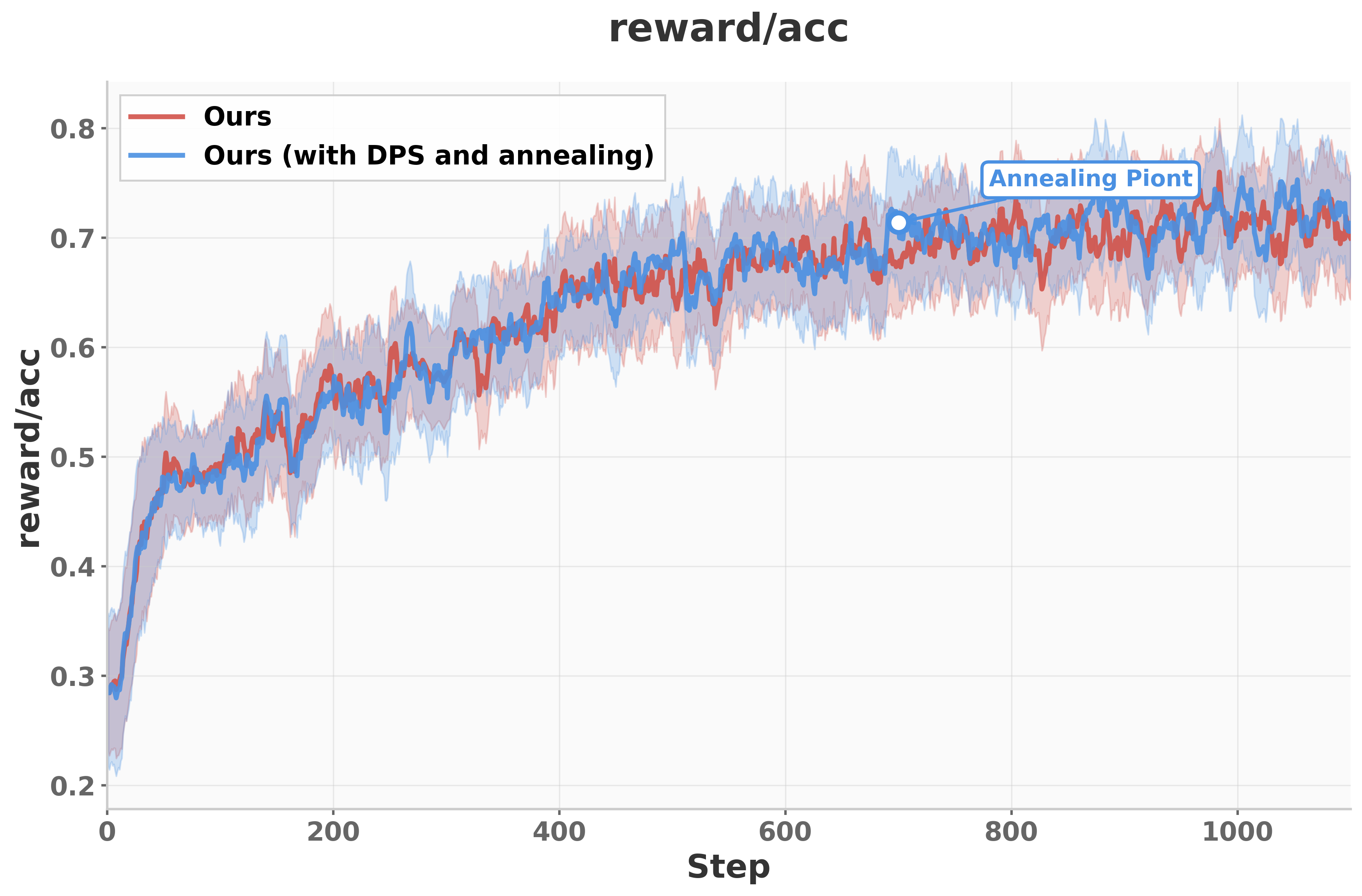}
    \caption{Reward comparison.}
    \label{reward}
  \end{subfigure}
  \hfill
  \begin{subfigure}[b]{0.32\linewidth}
    \includegraphics[width=\linewidth]{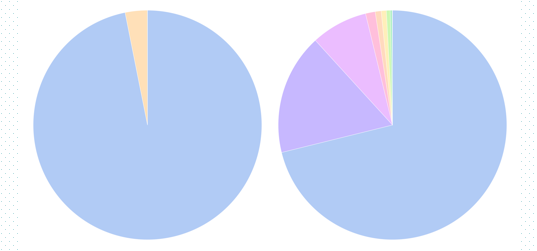}
    \caption{Trajectory diversity comparison.}
    \label{trajectory}
  \end{subfigure}

  \caption{Results about RQ5.}
  \label{para_4}
\end{figure*}

\textbf{RQ2: How does the model perform with different numbers of input images?}
To investigate our model's capability in processing varying numbers of images, we conduct experiments on MUIR (2-9 images) and MMIU (2-32 images) benchmarks. The experimental results are presented in Figures \ref{MUIR} and \ref{MMIU}. For different numbers of input images, our model outperforms the base model in most cases. Even when the number of input images exceeds 17, our model still achieves a significant improvement. This demonstrates the strong capability of our model in handling multi-image inputs and validates the effectiveness of the proposed cognition-inspired reasoning framework.

\textbf{RQ3: How does the model perform across different tasks?}
To explore the performance of our model across different tasks, we present the results on MUIR, which consists of 12 distinct tasks, in Figure \ref{task}. Our model achieves improvements on almost all tasks. Notably, tasks such as Geographic, Cartoon, and Visual Grounding were not included in our training set, yet our model still yields significant improvements on these tasks. This further demonstrates the generalization ability of our proposed reasoning framework in multi-image tasks.

\textbf{RQ4: Whether each meta-action plays a role in our framework?}
To demonstrate the contribution of each meta action in our framework, we conduct ablation studies by individually removing \texttt{<global>}, \texttt{<focus>}, \texttt{<hint>}, and \texttt{<think>} from CINEMA, and evaluate the performance on MUIR, MIRB, and VideoMME (with the performance showing in Table \ref{meta}). The results show that removing any of the actions leads to performance degradation, indicating that each meta-action plays an essential role in CINEMA.
\begin{table}[thb!]
  \centering
  \begin{tabular}{lccc}
    \toprule
     \textbf{Method} & \textbf{MUIR} & \textbf{MIRB} &\textbf{VideoMME}  \\
    \midrule
    Ours & \textbf{71.0} & \textbf{55.7} & \textbf{61.0}  \\
    \quad w/o \texttt{global} & 63.4 &52.6 & 57.1  \\
    \quad w/o \texttt{focus} & 61.6 &53.2 & 57.5  \\
    \quad w/o \texttt{hint}   & 63.5 &52.3 & 56.8 \\
    \quad w/o \texttt{think} & 60.1 &53.6 & 57.1 \\
    \bottomrule
  \end{tabular}
  \caption{Ablation study on meta actions.}
  \label{meta}
\end{table}

\textbf{RQ5: How does two-stage RL training influence entropy and training dynamics?}
Figure \ref{loss} and \ref{reward} illustrate the effect of two-stage RL training on entropy and learning dynamics. In the first stage, diversity-preserving strategy maintains moderate entropy, which then gradually declines during the second annealing stage. Compared to the baseline, our method consistently preserves higher entropy levels, effectively averting entropy collapse. This sustained entropy promotes continued exploration, prevents premature convergence, and helps retain a diverse set of meta-actions. Notably, despite the higher entropy, our approach matches the baseline in training accuracy, confirming that the entropy-preserving mechanism does not compromise performance. Figure \ref{trajectory} visualizes the trajectories, where each color denotes a distinct trajectory type. The right plot (with two-stage RL) exhibits richer and more varied meta-action trajectories than the left (without), demonstrating that our method sustains policy diversity even after annealing. The improved performance is further validated by the Pass@K results in Figure \ref{fig:pass}.

%% file: sec/6_conclusion.tex
\section{Conclusions}
In this work, we introduce CINEMA, a cognition-inspired meta-action framework that systematically decomposes multi-image reasoning into five structured actions: Global, Focus, Hint, Think, and Answer. By leveraging Retrieval-Based Tree Sampling for cold-start training and a two-stage reinforcement learning paradigm with diversity-preserving strategy and annealed DAPO, our approach effectively improves multi-image reasoning ability. Extensive experiments across multi-image, video, and single-image benchmarks demonstrate that CINEMA not only achieves state-of-the-art performance, surpassing even large general-purpose models such as GPT-4o in some key benchmarks, but also maintains higher policy diversity and adaptability. These results highlight the effectiveness, scalability, and generalizability of our framework, paving the way toward more robust multimodal reasoning systems.

%% file: sec/7_Acknowledgment.tex
\section*{Acknowledgment}
This research is funded by the National Nature Science Foundation of China (No.62477010, No.62577022 and No.62307028), AI for Science Program, Shanghai Municipal Commission of Economy and Informatization under Grant (2025-GZL-RGZN-BTBX-02018), the Natural Science Foundation of Shanghai (No.23ZR1441800), Shanghai Science and Technology Innovation Action Plan (No.24YF2710100 and No.23YF1426100),Shanghai Qiji Zhifeng Co., Ltd. (2025-GZL-RGZN-01001) and the opening funding of the State Key Laboratory of Disaster Reduction in Civil Engineering (Grant No.SLDRCE24-03).

%% file: sec/X_suppl.tex
\clearpage
\setcounter{page}{1}
\maketitlesupplementary
 
\section{Background: DAPO}
\label{app:DAPO}
DAPO \cite{yu2025dapo} is an improved variant of GRPO \cite{guo2025deepseek}, which directly computes the advantage $A_t$ using the average reward over multiple sampled outputs, thereby eliminating the need for a separate value function as in PPO. Specifically, given a prompt $\mathbf{q} \sim P(Q)$, we sample $G$ rollouts $\{\mathbf{o}_i\}_{i=1}^G$ from the current policy $\pi_{\theta_{\text{old}}}$. At each token position $t$ in rollout $i$, the likelihood ratio is defined in Eq.~\ref{eq:ration}.
\begin{equation}
  \small
\label{eq:ration}
r_{i,t}(\theta)
= \frac{\pi_{\theta}\bigl(o_{i,t}\mid \mathbf{q},\, \mathbf{o}_{i,<t}\bigr)}
       {\pi_{\theta_{\mathrm{old}}}\bigl(o_{i,t}\mid \mathbf{q},\,\mathbf{o}_{i,<t}\bigr)}
\end{equation}
The group-relative advantage \(\hat{A}_{i,t}\) is then obtained by standardizing each return \(R_i\) within the group, defined in Eq.~\ref{eq:advantage}.
\begin{equation}
  \small
\label{eq:advantage}
\hat{A}_{i,t}
= \frac{R_i \;-\;\mathrm{Mean}\bigl(\{R_j\}_{j=1}^G\bigr)}
       {\mathrm{Std}\bigl(\{R_j\}_{j=1}^G\bigr)}.
\end{equation}
In contrast to GRPO, DAPO introduces several methodological advancements. Specifically, it employs a Clip-Higher mechanism, wherein $\epsilon_{\text{high}}$ is set greater than $\epsilon_{\text{low}}$ to enhance exploratory behavior; integrates Dynamic Sampling to systematically exclude data instances lacking informative learning signals; incorporates an Overlong Punishment strategy to constrain excessively verbose outputs; and adopts a Token-level Loss formulation to mitigate the inherent bias between responses of varying lengths.
The training then proceeds by maximizing the clipped surrogate objective, defined for DAPO as follows:

\begin{equation} \label{eq:dapo}
\begin{aligned}
\mathcal{J}_{\text{DAPO}}(\theta) 
&= \mathbb{E}_{(q,a)\sim\mathcal{D},\,\{o_i\}_{i=1}^G \sim \pi_{\theta_{\text{old}}}(\cdot|q)}\\
\Biggl[ \frac{1}{\sum_{i=1}^{G} |o_i|} 
& \sum_{i=1}^{G}\sum_{t=1}^{|o_i|} \min\biggl( r_{i,t}(\theta)\hat{A}_{i,t}, \\
& \quad \operatorname{clip}\!\bigl(r_{i,t}(\theta),1-\varepsilon_{\text{low}},1+\varepsilon_{\text{high}}\bigr)\hat{A}_{i,t} \biggr) \Biggr], \\
\text{s.t.}\quad & 0 < \bigl|\{o_i \mid \text{is\_equivalent}(R_i,1)\}\bigr| < G.
\end{aligned}
\end{equation}

\section{Benchmark}
This section provides a detailed description of the benchmark used for evaluation.
\paragraph{MUIR} MUIRBENCH \citep{MUIR} is a comprehensive benchmark designed for robustly evaluating MLLMs’ multi-image understanding capabilities. It comprises 11,264 images and 2,600 multiple-choice questions (average 4.3 images per instance), covering 12 diverse multi-image tasks (e.g., action understanding, cartoon storytelling, geographic map reasoning, 3D object multiview retrieval).

\paragraph{MMIU}The Multimodal Multi-image Understanding (MMIU) \citep{MMIU} is a comprehensive benchmark tailored for evaluating MLLMs on multi-image comprehension tasks. Structured around cognitive psychology, it enumerates 7 types of multi-image relationships (refined from semantic, temporal, spatial categories) and covers 52 diverse tasks (e.g., multi-view action recognition, 3D object detection) .
In terms of scale, MMIU includes 77,659 images (2–32 per instance, averaging 6.64) and 11,698 meticulously curated multiple-choice questions.

\paragraph{MV-MATH} MV-MATH \citep{mvmath} is a specialized benchmark designed to evaluate MLLMs on mathematical reasoning in multi-visual contexts—addressing the gap in existing benchmarks that mostly focus on single images. It comprises 2,009 high-quality mathematical problems derived from real K-12 scenarios.

\paragraph{EMMA} EMMA \citep{EMMA} is a benchmark designed to evaluate Multimodal LLMs on genuine cross-modal reasoning. Its 2,788 questions across math, physics, chemistry, and coding require integrated visual-textual understanding, preventing solutions based on shallow cues or text alone.

\paragraph{Mantis-Eval} Mantis-Eval \citep{Mantis} is a benchmark dataset designed to evaluate a model's ability to reason across multiple images. It contains 217 challenging examples.

\paragraph{MIRB} MIRB \citep{MIRB} is a dedicated dataset addressing the gap in evaluating vision-language models (VLMs) on multi-image understanding, as existing benchmarks focus primarily on single-image inputs. It encompasses 925 samples across four core dimensions: perception, visual world knowledge, reasoning, and multi-hop reasoning, with all tasks requiring cross-comparison of multiple images (ranging from 2 to 42, averaging 3.78 per question).

\paragraph{MVBench} MVBench \citep{MVBench} is a multi-modal video benchmark addressing the lack of temporal understanding evaluation in MLLMs, covering 20 multi-frame-dependent video tasks (defined via a static-to-dynamic method). It is built efficiently by auto-converting public video annotations into multiple-choice QA (with ground-truth for fairness), reveals existing MLLMs’ poor temporal understanding.

\paragraph{Video-MME} Video-MME \citep{VideoMME} is the first comprehensive benchmark designed to evaluate MLLMs in video analysis. It fills the gap in assessing the understanding of sequential visual data by featuring 900 videos (ranging from 11 seconds to 1 hour) across 6 core domains (e.g., Knowledge, Sports Competition) and 30 subfields. Each video is paired with three expert-annotated multiple-choice QA pairs, resulting in a total of 2,700 questions. To support multi-modal reasoning, the benchmark also provides subtitles for 744 videos and audio tracks for all 900 videos.

\paragraph{Video-MMMU} Video-MMMU \citep{VideoMMMU} is a benchmark designed to evaluate the knowledge acquisition capabilities of MLLMs from professional video content. It comprises 300 expert-level videos spanning six disciplines  and 30 subfields, paired with 900 human-annotated question–answer pairs. The benchmark measures performance across three cognitive stages: (1) \emph{Perception}, assessing whether models can extract salient knowledge-related details from video content; (2) \emph{Comprehension}, evaluating the ability to grasp and reason about the underlying concepts; and (3) \emph{Adaptation}, examining whether models can transfer the acquired knowledge to novel or unfamiliar scenarios.

\paragraph{MMMU-Pro} MMMU-Pro \citep{MMMU-Pro} is an enhanced version of the MMMU benchmark, designed to more rigorously evaluate multimodal models’ understanding and reasoning.

\paragraph{M3CoT} M3CoT \citep{m3cot} addresses gaps in existing MCoT benchmarks (lack of visual reliance, single-step reasoning, limited domains) by enabling multi-domain, multi-step, multi-modal reasoning across 3 domains (science, mathematics, commonsense), 17 topics, and 263 categories. It has 11,459 total samples (7,973 train, 1,127 dev, 2,359 test) with diverse image types (geographic graphs, health images, etc.).

\paragraph{MM-MATH} MM-MATH\citep{MM-MATH}consists of 5,929 open-ended middle school math problems paired with visual contexts, and it adopts fine-grained classification covering three dimensions: difficulty, grade level, and knowledge points.

\paragraph{MathVista} MathVista \citep{Mathvista} is proposed as a benchmark integrating challenges from mathematical and visual tasks. It contains 6,141 examples, sourced from 28 existing multimodal math datasets and 3 new ones (IQTest, FunctionQA, PaperQA), requiring fine-grained visual understanding and compositional reasoning—tasks that state-of-the-art foundation models find challenging.

\paragraph{MATH-V} MATH-V \citep{Math-Vsion} is a curated dataset designed to address the limited question diversity and subject breadth of existing visual math reasoning benchmarks. It comprises 3,040 high-quality math problems with visual contexts, all sourced from real math competitions.

\section{Training Data Construction}
\label{construction}

For the construction of the training dataset, we referenced Mantis \citep{Mantis}, LLaVA-Interleave \citep{Llava-next-interleave}, Leopard \cite{Leopard}, and VideoR1 \cite{Videor1}. Overall, our dataset consists of multi-image data and single-image data, with 57k samples for cold-start training and 58k samples for RL. The detailed dataset statistics are presented in Table 1. Regarding the partitioning criteria for RL data and cold-start data, the key distinction between our cold-start and reinforcement learning dataset splits lies in the trajectory generation process described in Section \ref{sec3.2}. Cold-start training data consists of problems where GPT-4o successfully provides correct answers during Step 2, and for these instances, we proceed to Step 3 (retrieval-based diverse sampling) to obtain two distinct correct reasoning trajectories per problem that serve as supervised learning targets for cold start training. In contrast, reinforcement learning data comprises problems where GPT-4o fails to produce correct answers during Step 2, and these challenging cases are reserved for reinforcement learning.

\begin{table*}[thb!]
\setlength{\tabcolsep}{2pt}
  \centering
 
  \begin{tabular}{cccc}
    \toprule
    Type&  Dataset & Count for SFT  & Count for RL \\
    \midrule
\multirow{10}{*}{Multi-Image}  & ChartVQA\citep{Leopard}  & 2501 & - \\
                            & SlideVQA\citep{Leopard}   & 3249 & 3000 \\
                            & ALFRED\citep{ALFRED}  &8357 & -\\
                            & Nuscenes\citep{Nuscenes}   &- &4946\\
                            & RecipeQA\citep{recipeqa} &8759 &5069 \\
                            & IconQA\citep{IconQA} &5315 &3000 \\
                            & nlvr2\citep{nlvr2} &5424 &1620 \\
                            & Spot-the-Diff\citep{spot} &2248 &2589 \\
                            & LRV\citep{LRV} &- &2993 \\
                            & RAVEN\citep{raven} &- & 3200\\

\midrule
\multirow{5}{*}{Video}  & Star\citep{star}  & 5490 & 2754 \\
                            & NextQA\citep{nextqa}   & 1193 & 3000 \\
                            & Clevrer\citep{clevrer_video}  &3047 &4478\\
                            & Perception\citep{perception}   &2964 &2500\\
\midrule
\multirow{15}{*}{Single-Image}  
                            & Clevr\_cogen\_a\_train\tablefootnote{\url{https://huggingface.co/datasets/leonardPKU/clevr_cogen_a_train}}  & 1506 & - \\
                            & Clevr\_CoGenT\_TrainA\_70K\_Complex\tablefootnote{\url{https://huggingface.co/datasets/MMInstruction/Clevr_CoGenT_TrainA_70K_Complex}}    & 1159 & 3000 \\
                            & M3COT\citep{m3cot}  &1147 &-\\
                            & Share-GRPO\citep{shareGRPO}   &1145 &3000\\
                            & GEOQA\_R1V\_Train\_8K\tablefootnote{\url{https://huggingface.co/datasets/leonardPKU/GEOQA_R1V_Train_8K}}    &800 &4816\\
                            & AI2D\citep{ai2d}   &630 &-\\
                            & MMK12\citep{MMMK12}   &442 &3537\\
                            & Geometry3k\citep{geometry3k}   &317 &1406\\
                            & ScienceQA\citep{ScienceQA}   &259 &-\\
                            & PISC\citep{PISC}   &244 &-\\
                            & Geoqa+\citep{geoqa+}   &172 &891\\
                            & GQA\citep{gqa}   &119 &-\\
                            & CLEVR\_v1.0\citep{clevr}   &118 &-\\
                            & COCO\citep{coco}   &78 &-\\
                            & LRV\citep{LRV}   &- &3063\\

    \bottomrule
  \end{tabular}
   \caption{Statistics of Training Data}
  \label{tab:MME}
\end{table*}

\section{Implementation}
\label{Implementation}
Our SFT experiments are primarily conducted using the LLaMA Factory framework \citep{Zheng_LlamaFactory_Unified_Efficient_2024}, with the main hyperparameters summarized in Table \ref{SFT_hyper}. For the RL stage, we rely on the EasyR1 framework \cite{zheng2025easyr1}, a multi-model large-scale training system built upon VERL \citep{sheng2024hybridflow}, and the key parameters are reported in Table \ref{RL_hyper}. The experiments run on 32 A800 GPUs.

\section{Case Study}
\begin{figure*}[!htb]
    \begin{subfigure}[b]{\textwidth}
        \centering
        \includegraphics[width=0.8\textwidth]{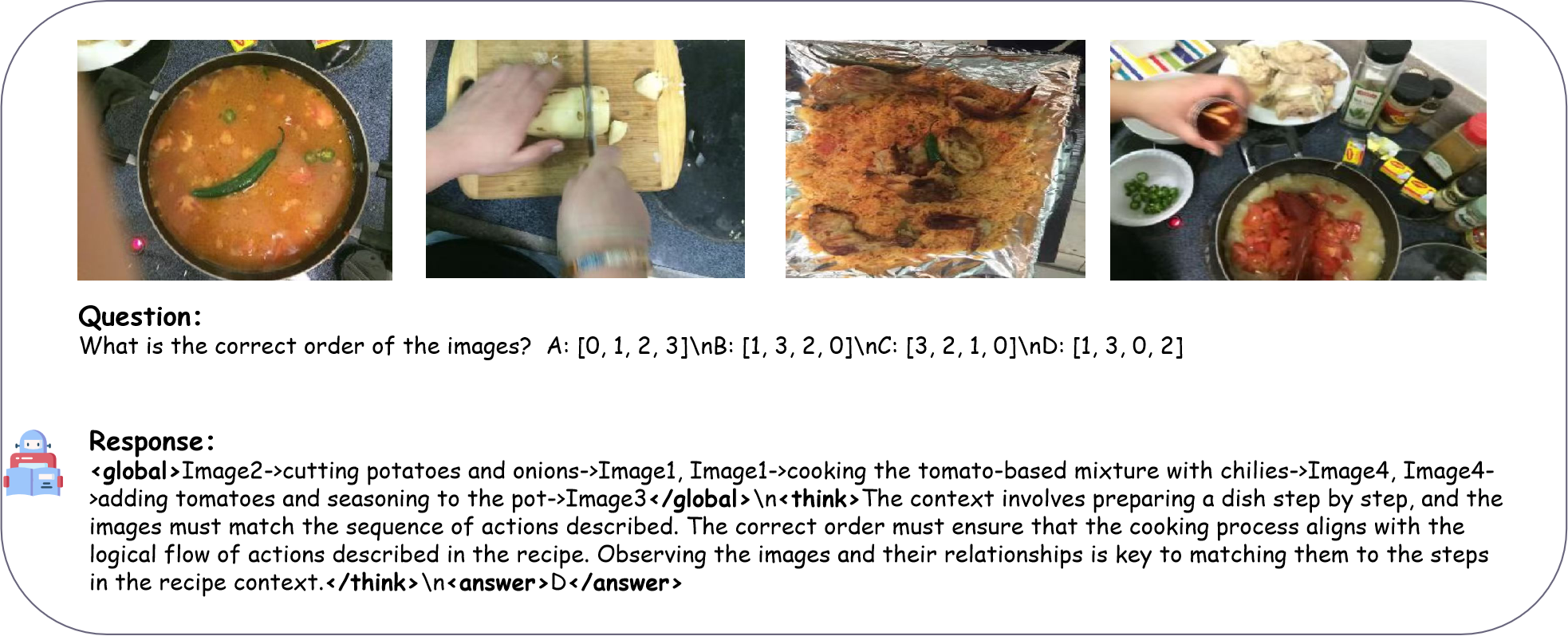}
        \caption{Case1.}
        \label{fig:img1}
    \end{subfigure}
    \vspace{1cm}
    
    \begin{subfigure}[b]{\textwidth}
        \centering
        \includegraphics[width=0.8\textwidth]{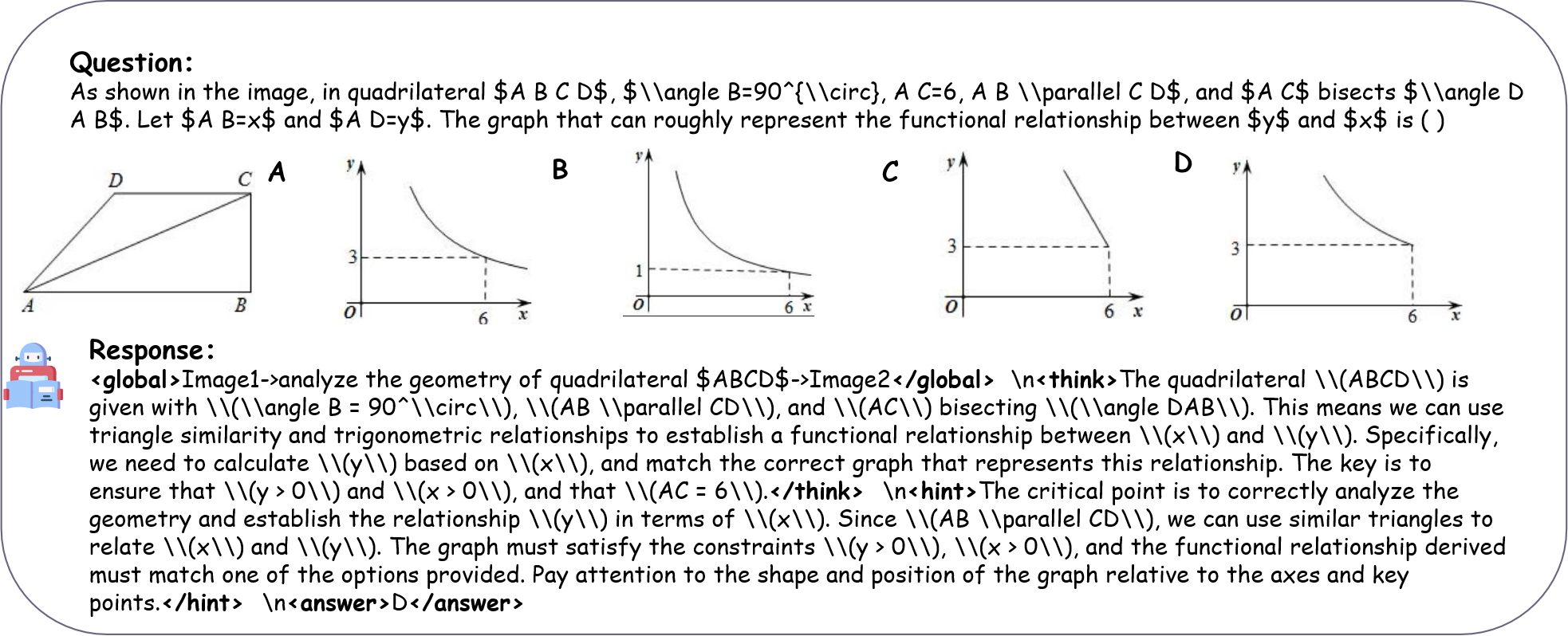}
        \caption{Case2.}
        \label{fig:img4}
    \end{subfigure}
    \vspace{1cm}

    \begin{subfigure}[b]{\textwidth}
        \centering
        \includegraphics[width=0.8\textwidth]{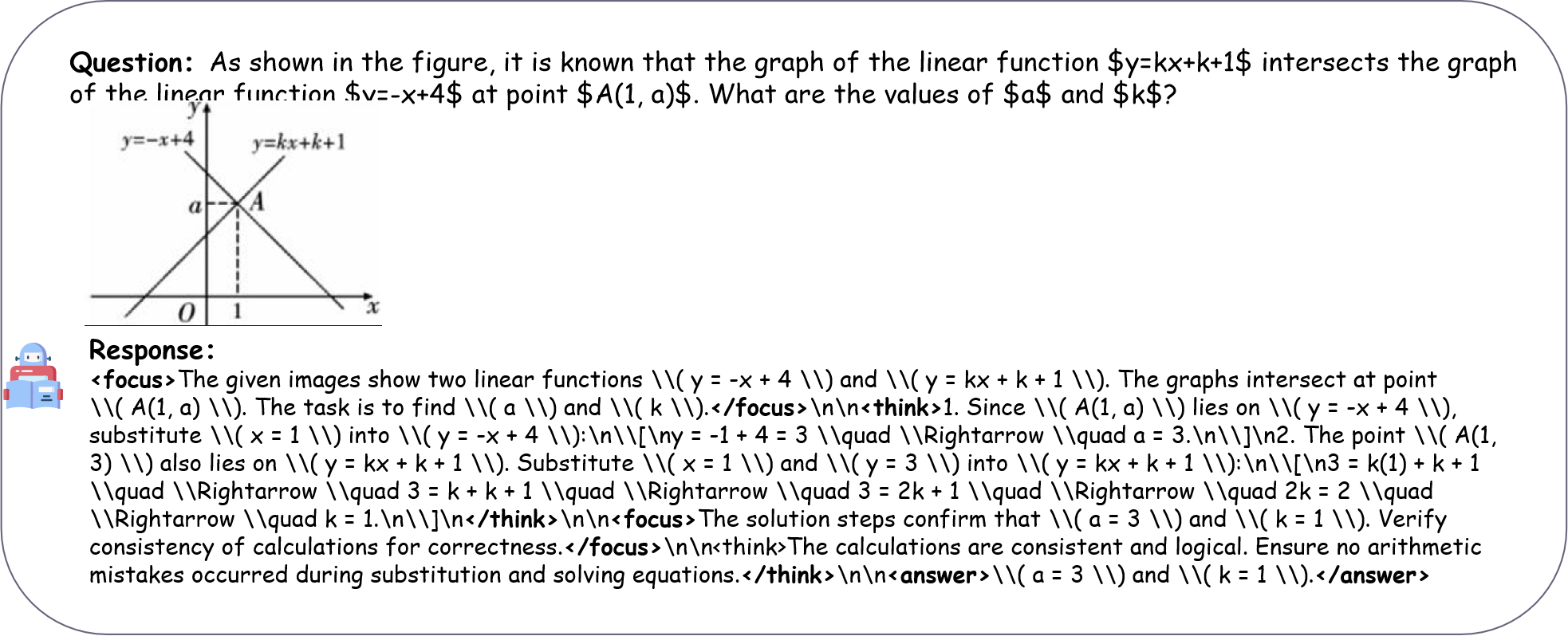}
        \caption{Case3.}
        \label{fig:img5}
    \end{subfigure}
    
    \caption{Case study.}
    \label{fig:all_images}
\end{figure*}

\begin{figure*}[!htb] 

    \centering

    \begin{subfigure}[b]{\textwidth}
        \centering
        \includegraphics[width=0.8\textwidth]{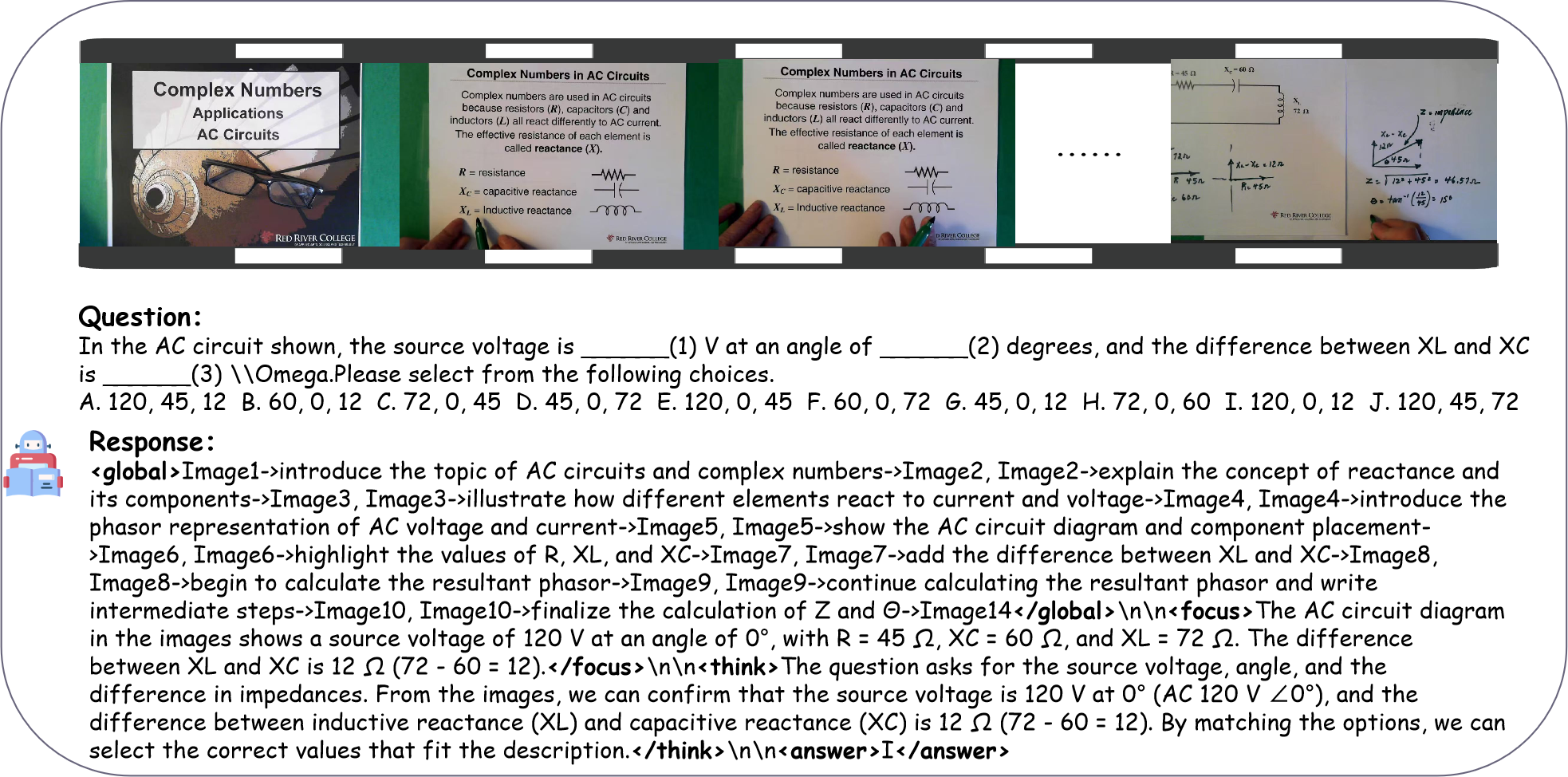}
        \caption{Case4.}
        \label{fig:img2}
    \end{subfigure}
    \vspace{1cm}

    \begin{subfigure}[b]{\textwidth}
        \centering
        \includegraphics[width=0.8\textwidth]{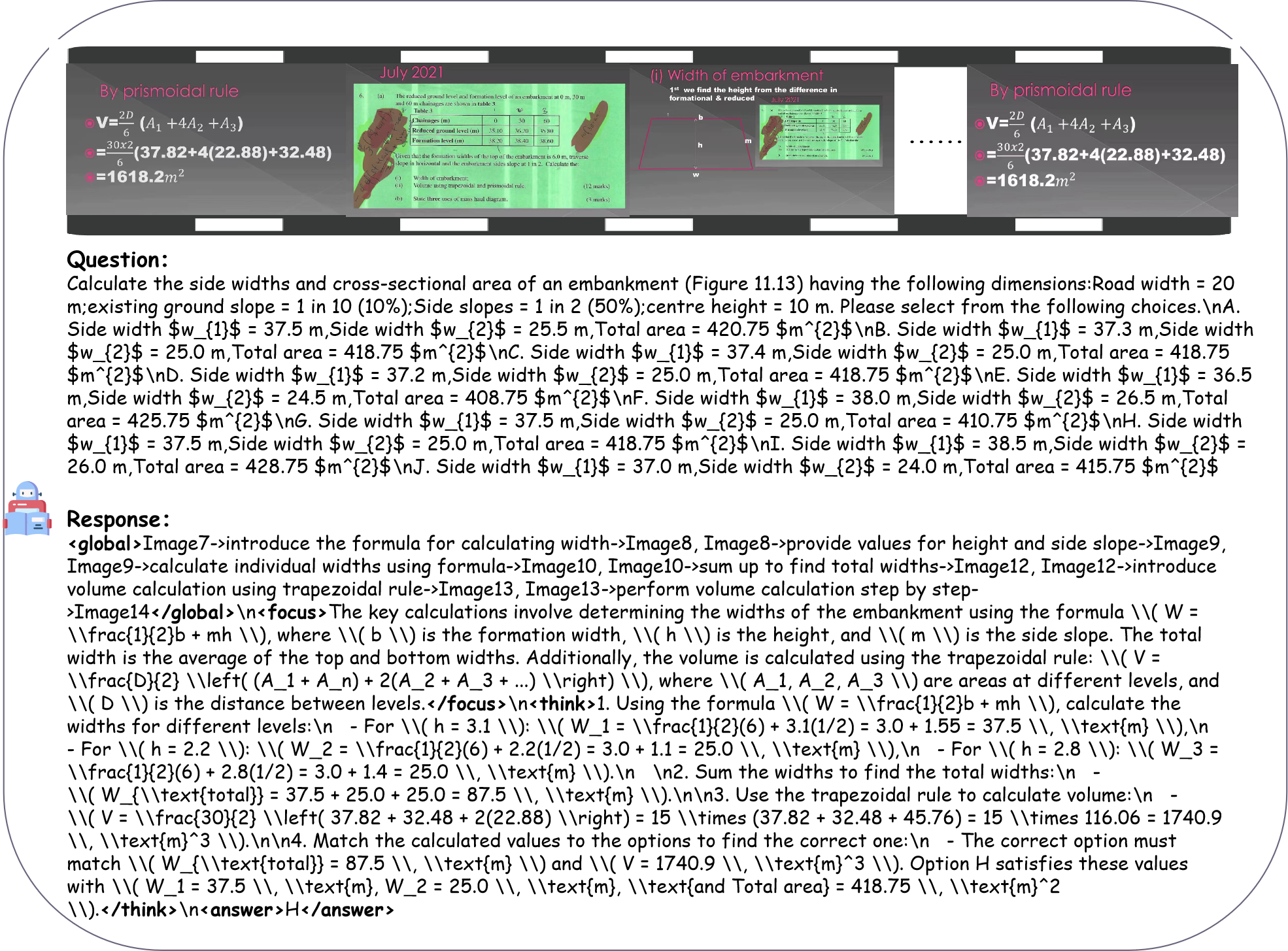}
        \caption{Case5.}
        \label{fig:img3}
    \end{subfigure}
    \caption{Case study.}
    \label{fig:all_images_part1}
\end{figure*}

\begin{figure*}[]
  \centering
  \includegraphics[width=0.8\textwidth]{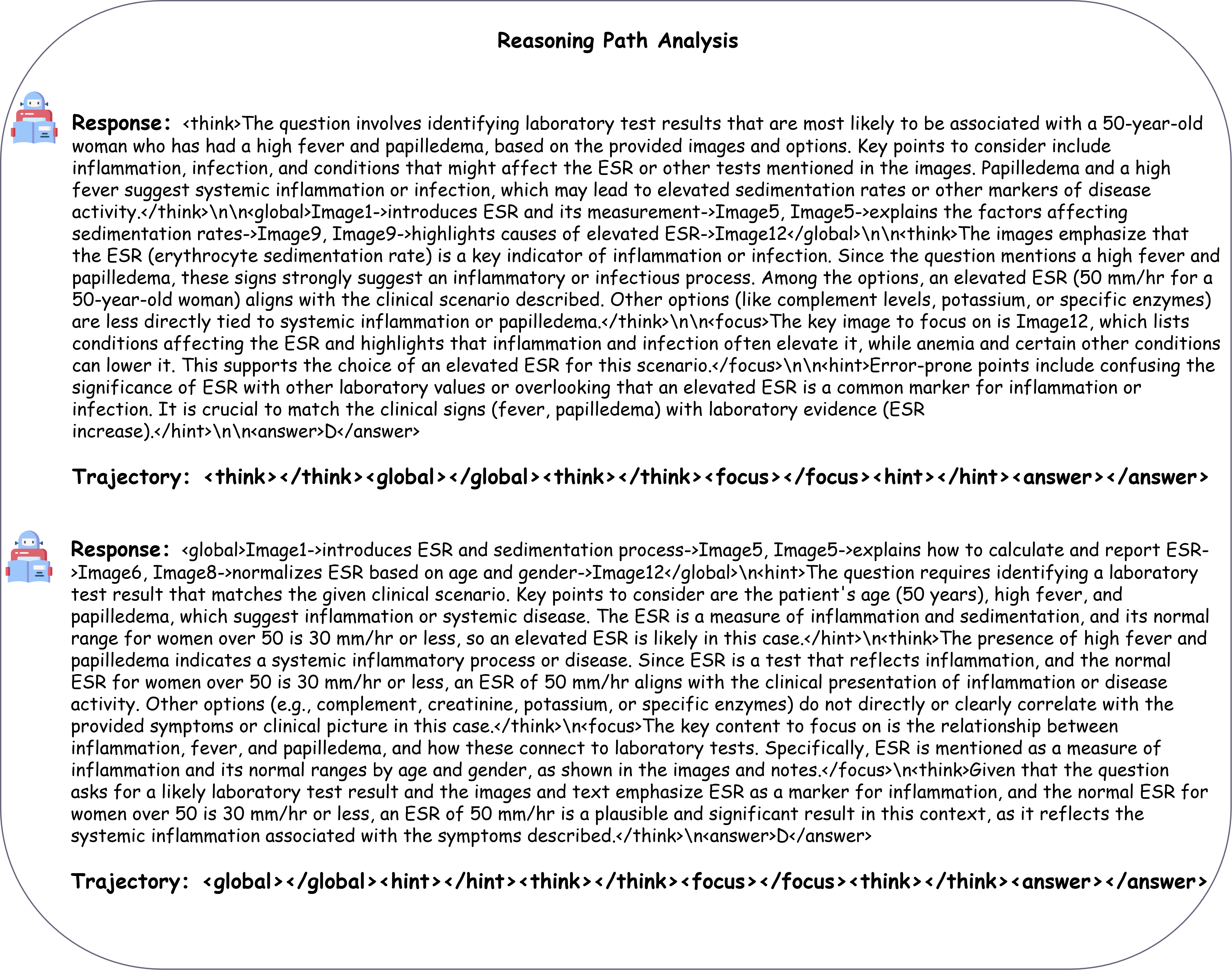}\\
  \caption{Reasoning Path Analysis.}
  \label{case_reaosning_path}
\end{figure*}

Here we present a case study of our model in Figure \ref{fig:all_images} and \ref{fig:all_images_part1}, covering multi-image benchmarks, video benchmarks, and single-image benchmarks. The results demonstrate that, across different types of tasks, our model can dynamically invoke appropriate meta-actions to analyze the problem and produce correct answers. In Figure \ref{case_reaosning_path}, we present multiple reasoning results for a single problem. It can be observed that the model explores different reasoning paths for the same problem, all of which lead to the correct answer.

\begin{table}[htbp]
\centering
\begin{tabular}{ll}
\toprule
\textbf{Parameter} & \textbf{Value} \\
\midrule

\multicolumn{2}{l}{\textbf{Model}} \\
model\_name\_or\_path & Qwen2.5-VL-7B-Instruct \\
image\_max\_pixels & 100352 \\
\midrule

\multicolumn{2}{l}{\textbf{Method}} \\
stage & sft \\
do\_train & true \\
finetuning\_type & full \\
\midrule

\multicolumn{2}{l}{\textbf{Dataset}} \\
template & qwen2\_vl \\
cutoff\_len & 12000 \\
overwrite\_cache & true \\
preprocessing\_num\_workers & 16 \\
dataloader\_num\_workers & 4 \\
\midrule

\multicolumn{2}{l}{\textbf{Train}} \\
per\_device\_train\_batch\_size & 1 \\
gradient\_accumulation\_steps & 4 \\
learning\_rate & 1.0e-5 \\
num\_train\_epochs & 2 \\
lr\_scheduler\_type & cosine \\
warmup\_ratio & 0.1 \\

\bottomrule
\end{tabular}
\caption{Hyperparameters used in SFT.}
\label{SFT_hyper}
\end{table}

\begin{table}[htbp]
\centering
\begin{tabular}{ll}
\toprule
\textbf{Parameter} & \textbf{Value} \\
\midrule

\multicolumn{2}{l}{\textbf{Data}} \\
max\_prompt\_length & 4096 \\
max\_response\_length & 4096 \\
rollout\_batch\_size & 64 \\
max\_pixels & 100352 \\
min\_pixels & 50176 \\
\midrule

\multicolumn{2}{l}{\textbf{Algorithm}} \\
adv\_estimator & grpo \\
kl\_coef & 0.0 \\
filter\_groups\_enable & true \\
filter\_max\_num\_gen\_batches & 20 \\
filter\_metric & acc \\
\midrule

\multicolumn{2}{l}{\textbf{Worker.Actor}} \\
global\_batch\_size & 32 \\
max\_grad\_norm & 1.0 \\
entropy\_coeff & 0.0 \\
kl\_loss\_coef & 0.0 \\
clip\_ratio\_low & 0.2 \\
clip\_ratio\_high & 0.28 \\
optim.lr & 1.0e-6 \\
optim.weight\_decay & 1.0e-2 \\
\midrule

\multicolumn{2}{l}{\textbf{Worker.Rollout}} \\
temperature & 1.0 \\
top\_p & 1.0 \\
top\_k & -1 \\
n & 8 \\

\bottomrule
\end{tabular}
\caption{Hyperparameters used in RL.}
\label{RL_hyper}
\end{table}